\documentclass{article}

\usepackage{microtype}
\usepackage{graphicx}
\usepackage{subfigure}
\usepackage{booktabs} 

\usepackage{amsmath}
\usepackage{amsthm}
\usepackage{amsfonts}
\usepackage{amssymb}

\usepackage{mathtools, cuted}

\newcommand{\OO}{O} 

\newcommand{\X}{X}

\newcommand{\PB}{p_{\text{binom}}}

\usepackage{algorithm,algorithmic}
\usepackage{wrapfig}
\renewcommand{\algorithmicrequire}{\textbf{Input:}}
\renewcommand{\algorithmicensure}{\textbf{Output:}}

\newcommand{\NBATCH}{{K}}

\newtheorem{lemma}{Lemma}
\newtheorem{theorem}{Theorem}
\newtheorem{assumption}{Assumption}
\newtheorem{definition}{Definition}
\newtheorem{corollary}{Corollary}

\usepackage{hyperref}

\usepackage[accepted]{icml2022}

\icmltitlerunning{Analysis of Stochastic Processes through Replay Buffers}

\renewcommand{\X}{X}
\newcommand{\RB}{RB}

\begin{document}

\twocolumn[
\icmltitle{Analysis of Stochastic Processes through Replay Buffers}



\icmlsetsymbol{equal}{*}

\begin{icmlauthorlist}
\icmlauthor{Shirli Di~Castro Shashua}{technion}
\icmlauthor{Shie Mannor}{technion,nvidia}
\icmlauthor{Dotan Di~Castro}{bcai}
\end{icmlauthorlist}

\icmlaffiliation{technion}{Technion Institute of Technology, Haifa, Israel}
\icmlaffiliation{nvidia}{NVIDIA Research, Israel}
\icmlaffiliation{bcai}{Bosch Center of AI, Haifa, Israel}

\icmlcorrespondingauthor{Shirli Di~Castro Shashua}{sdicastro@gmail.com}

\icmlkeywords{Machine Learning, ICML}

\vskip 0.3in
]



\printAffiliations{}

\begin{abstract}
Replay buffers are a key component in many reinforcement learning schemes. Yet, their theoretical properties are not fully understood. In this paper we analyze a system where a stochastic process $X$ is pushed into a replay buffer and then randomly sampled to generate a stochastic process $Y$ from the replay buffer. We provide an analysis of the properties of the sampled process such as stationarity, Markovity and autocorrelation in terms of the properties of the original process. Our theoretical analysis sheds light on why replay buffer may be a good de-correlator. Our analysis provides theoretical tools for proving the convergence of replay buffer based algorithms which are prevalent in reinforcement learning schemes.
\end{abstract}

\section{Introduction}
A Replay buffer (RB) is a mechanism for saving past generated data samples and for sampling data for off-policy reinforcement learning (RL) algorithms \cite{lin1993reinforcement}. The RB serves a First-In-First-Out (FIFO) buffer with a fixed capacity and it enables sampling mini-batches from previously saved data points. Its structure and sampling mechanism provide a unique characteristic: the RB serves as \textit{de-correlator} of data samples. Typically, the agent in RL algorithms encounters sequences of highly correlated states and  learning from these correlated data points may be problematic since many deep learning algorithms suffer from high estimation variance when  data samples are dependent. Thus, a mechanism that decorrelates the input such as the RB can improve data efficiency and reduce sample complexity.

Since its usage in the DQN algorithm \cite{mnih2013playing},  RB mechanism have become popular in many off-policy RL algorithms \cite{lillicrap2015continuous,haarnoja2018soft}. Previous work has been done on the empirical benefits of RB usage \cite{fedus2020revisiting,zhang2017deeper}, but still there is a lack in \text{theoretical} understanding of how the RB mechanism works. Understanding the properties of RBs is crucial  for convergence and finite sample analysis of algorithms that use a RB in training. For the best of our knowledge, this is the first work to tackle these theoretical aspects.

In this work we focus on the following setup. We define a random process $X$ that is pushed into a $N$ samples size RB and analyze the characteristics of the stochastic process of $K$ samples that is sampled from the RB. We analyze if properties of the original random process such as Markovity and stationarity are maintained and quantify the  auto-correlation and covariance in the new RB process (later denoted by $Y$) when possible. 

Our motivation comes from RL algorithms that use RB. Specifically, we focus on the induced Markov chain given a policy but we note that the analysis in this paper is also relevant to general random processes that are kept in a FIFO queue. This is relevant for domains such as First Come First Served domains \cite{laguna2013business}. Our goal is to provide analytical tools for analyzing algorithms that use RBs. Our results can provide theoretical understanding of phenomena seen in experiments using RBs that have never been analyzed theoretically before. Our  theory for RBs provides tools for proving convergence of RB-based RL algorithms. 
\begin{figure*}[t]
\centerline{\includegraphics[width=1.7\columnwidth]{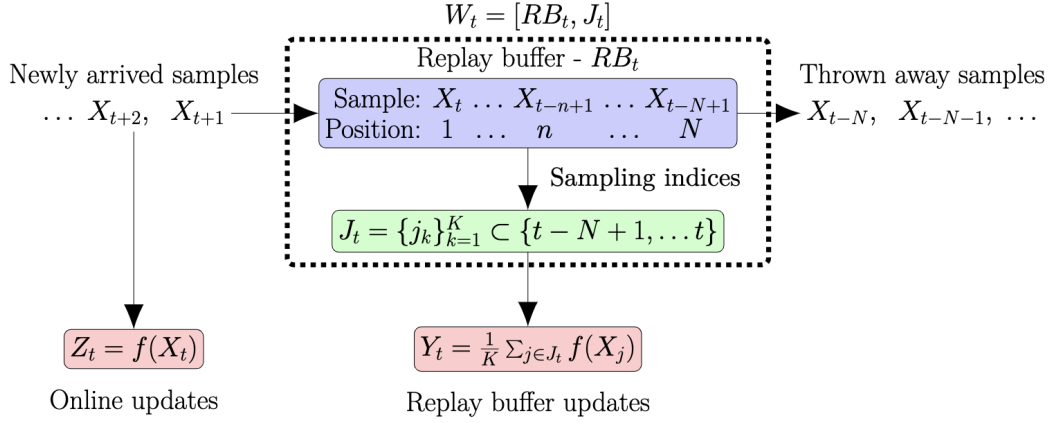}}
\caption{Replay buffer flow diagram: Process $X$ enters the RB which stores $\{X_{t}, \ldots X_{t-N+1}\}$ in positions $(1, \ldots, N)$, respectively. As time proceeds and $t>N$, old transitions are thrown away from the RB. At each time step $t$, a random subset $J_t$ of $K$ time steps is sampled from the RB. $W$ is simply  $[RB, J]$. $Y$ is the process of averaging a function over X at times from the subset $J$. Lastly, the process Z is simply a function applied on the variable X. Comparing Y to Z, we can see that Z can serve as an online update while Y can serve as a RB-based update.}
\label{fig:RB_diagram}
\end{figure*}
Our main contributions are:
\begin{enumerate}
    \item Formulating RBs as random processes and analyze their properties such as stationarity, Markovity, ergodicity, auto-correlation and covariance.
    \item Comparing between properties of the original random process that was pushed into the RB and the sampled process at the output of the RB. Particularly we prove that when sampling uniformly from the RB, the RB forms as a de-correlator between the sampled batches. 
    \item Connecting our RB theory to RL by demonstrating this connection through a RB-based actor critic RL algorithm that samples $K$ transitions from RB with size $N$ for updating its parameters. We prove, for the first time, the asymptotic convergence of such RB-based actor critic  algorithm.
\end{enumerate}

The paper is structured as follows. We begin with presenting the setup in Section \ref{section:setup}. We then state our main results regarding RB properties in Section \ref{section:RBProperties}. In Section  \ref{section:RB_in_RL} we connect between our RB theory and its use in RL and provide a convergence proof for an RB-based actor critic algorithm. Afterward, in Section \ref{section:related_work} we position our work in existing literature and conclude in Section \ref{section:conclusions}.

\section{Setup for Replay Buffer Analysis}
\label{section:setup}

\subsection{Replay Buffer Structure}
\label{sec:RB_structure}
Let $\X \triangleq (\X_t)_{t=0}^{\infty}$ be a stochastic process where the subscript $t$ indicates time. The samples are dynamically pushed into a Replay Buffer (RB;  \citealp{lin1993reinforcement,mnih2013playing}) of capacity $N$, i.e.,  it is a First-In-First-Out (FIFO) buffer that can hold the $N$ latest samples. We define the state of the RB at time $t$ with $\RB_t = \{\X_{t-N+1}, \ldots, \X_t \}$. Suppose that the buffer cells are numbered from $1$ to $N$. The latest observation of $\X$ is pushed into cell $1$, the observation before into cell $2$, etc. When a new observation arrives, the observation in cell $n$ is pushed into cell $n+1$ for $1 \le n < N$, while the observation in cell $N$ is thrown away.

The random process $\RB=(\RB_t)_{t=0}^{\infty}$ contains the last $N$ samples of $X$. The random process $Y$ is defined as the average of random $K$  samples (without replacement) out of the $N$ samples and applying a function $f(\cdot):\cdot \rightarrow \mathbb{R}^D$ \footnote{We note that $f(\cdot)$ may also depend on $t$ but we leave that for the sake of simplicity.} where $D$ is the dimension of the algorithm\footnote{For example, in linear function approximation of Actor-Critic algorithms, $D$ is the dimension of the linear basis used to approximate the value function by the critic.}.  The function $f(\cdot)$ may correspond to a typical RL function that one usually find in RL algorithms such as linear function approximation, Temporal Difference, etc. (\citealp{bertsekas1996neuro}). We elaborate on possible RL functions in Section \ref{section:connect_RB_RL_functions}.  

\subsection{Replay Buffer Sampling Method}
\label{sec:smampling_method}
We analyze the "unordered sampling without replacement" strategy from the RB. We note that other sampling methods may be analyzed, but we chose this specific sampling due to its popularity in many deep reinforcement learning algorithms\footnote{In Section \ref{section:conclusions} we discuss shortly future directions for other sampling schemes.}. Let $\mathbb{N}$ be a set of indices: $\mathbb{N} = \{ 1, \ldots, N\}$ and let $\bar{J}$ be a subset of $K$ indices from $\mathbb{N}$. Given $N$ and $K$, let $\mathbb{C}_{N,\NBATCH}$ be the set of all possible subsets $\bar{J}$ for specific $N$ and $\NBATCH$. Then, the probability of sampling subset $\bar{J}$ is $\PB^{N,\NBATCH}(\bar{J}) = \frac{1}{\binom{N}{K}} \ \forall \bar{J} \in \mathbb{C}_{N,K}$, where $\binom{N}{K} \triangleq \frac{N!}{(N-K)! K!}$ is the binomial coefficient. 

\subsection{Replay Buffer Related Processes}

We denote the set of $K$ temporal indices of the samples from $\RB$ by the random process $J$ (corresponds to a \emph{"Batch"} in Deep Learning) where $J_t = \{j_k\}_{k=1}^K \subset \{t-N+1,\ldots,t\}$\footnote{We note that in the first $K$ steps the batch is of size smaller than $K$ and in the first $N$ steps the RB is not full.}. Similarly, the corresponding $K$ RB indices process is $\bar{J}$ where $\bar{J}_t \subset \{1,\ldots,N\}$. We remark that both $J_t$ and $\bar{J}_t$ contain identical information but one refers to the absolute time, and one to the indices of the RB. We define the random process $W_t\triangleq[\RB_t, J_t]$ which holds both the information on the RB as well on the sampling from it. For later usage, we define the process $X_t$ going through a function $f(\cdot)$ with $Z_t \triangleq f(X_t)$. The resulting $Y_t$ has the structure of 
\begin{equation*}
    Y_t = \frac{1}{K}\sum_{j \in J_t}Z_j = \frac{1}{K}\sum_{j \in J_t}f(X_j).
\end{equation*}
The stochastic processes relations that are described above are visualized in Figure \ref{fig:RB_diagram}. 

\section{Replay Buffer Properties}
\label{section:RBProperties}
In this section we analyze the properties of a random process $Y$ that is sampled from the RB and used in some RL algorithm. Specifically, we analyze stationarity, Markovity, ergodicity, auto-correlation, and covariance.

\subsection{Stationarity, Markovity and Ergodicity}
The following Lemmas characterize the connection between different properties of $X$ that enter the RB and the properties of the processes RB and Y.

Stationarity is not a typical desired RL property since we constantly thrive to improve the policy (and thus the induced policy) but we bring it here for the sake of completeness.
\begin{lemma}[Stationarity]
\label{lemma:Yt_is_stationary}
Let $X_t$ and $\bar{J}_t$ be stationary processes. Then, $\RB_t$ and $Y_t$ are stationary. 
\end{lemma}

The proof for Lemma \ref{lemma:Yt_is_stationary} is deferred to Section \ref{appendix:proof_stationarity} in the supplementary material. 

In the next Lemma we analyze when the process $\RB$ is Markovian. This property is important in RL analysis. 
Note that $Y_t$ is not necessarily Markovian, however, $W_t$ is Markovian. For this, with some abuse of notation, we define $\X_{n_1}^{n_2} (t) \triangleq \{\X_{t - n_2 + 1}, \ldots, \X_{t - n_1 + 1} | \RB_t\}$ as the set of random variables realizaions from process $\X$ stored in the RB at time $t$  from position $n_1$ to $n_2$. 

\begin{lemma}[Markovity]
\label{lemma:RB_is_markovian}
Let $X_t$ be a Markov process. Then: (1) $\RB_t$ and $W_t$ are Markovian. (2) The transition probabilities of $\RB_t$ for $t \ge N$ are: 
\begin{align*}
    & P \left(RB_{t+1} \middle| RB_{t} \right) \\
    & = \begin{cases}
P \left(X_{t+1}   \middle| \X_t \right) & \text{ if } X_t \in X_1^1(t) \\
& \text{ and } RB_{t+1} = \{X_{t+1}\} \cup X_1^{N-1}(t), \\
0  & \text{ otherwise. }
\end{cases}
\end{align*}

If $J_t$ is sampled according to "unordered sampling without replacement", then the transition probabilities of $W_t$ for $t \ge N$ are:

\begin{align*}
    & P \left( W_{t+1} \middle| W_t \right) \\
    & = \begin{cases}
\frac{1}{\binom{N}{K}} P \left(X_{t+1}   \middle| \X_t \right) & \forall J_{t+1} \in \mathbb{C}_{N, K}, \text{ if } X_t \in X_1^1(t) \\
& \text{ and } RB_{t+1} = \{X_{t+1}\} \cup X_1^{N-1}(t) \\
0  & \text{ otherwise. }
\end{cases}
\end{align*}
\end{lemma}
The proof for Lemma \ref{lemma:RB_is_markovian} is deferred to Section \ref{appendix:proof_RB_is_markovian} in the supplementary material.

In RL, the properties of aperiodicity and irreducible (that together form ergodicity; \citealp{norris1998markov}) are crucial in many convergence proofs. The following states that these properties are preserved when using RB. We denote $\text{supp}(Y_t)$ as the set of all the values $Y_t$ can take. 
\begin{lemma}[Ergodicity]
\label{lemma: ergodicity}
Let $X_t$ be a Markov process that is aperiodic and irreducible. Then, $\RB_t$ and $W_t$ are aperiodic and irreducible. Moreover, every point $y\in\textrm{supp}(Y_t)$ is visited infinitely often.
\end{lemma}

The proof for Lemma \ref{lemma: ergodicity} is deferred to Section \ref{appendix:proof_ergodicity} in the supplementary material.

\subsection{Auto-Correlation and Covariance}
In this section we analyze the auto-correlation and covariance of the process $Y$ expressed by process $X$ properties. When $X$ is stationary, the auto-correlation and covariance functions for $X$ are: 
\begin{align*}
    R_X(\tau) & = \mathbb{E}[X_t X_{t+\tau}] \\
    C_X(\tau) & = \mathbb{E} \left[ \left(X_{t} - \mathbb{E} \left[ X_{t} \right] \right) \left(X_{t+\tau} - \mathbb{E} \left[ X_{t+\tau} \right] \right) \right]
\end{align*}

In the same way, the definition of the auto-correlation and covariance functions for the process $Y$ are $R_Y(\tau)$ and $C_Y(\tau)$, respectively. In the following theorem we prove the relationship between the auto-correlation and covariance functions of processes $Y$ and $X$. For that, we need to define the distribution of all time differences between two batches of samples as follows.
\begin{definition}[Time difference distribution between two batches of samples]\label{definition:tau_tag}
Consider a RB of size $N$. Consider taking two different random permutations (batches), denoted by $\bar{J}_a$ and $\bar{J}_b$, both of length $\NBATCH$ in two possibly different time points, $t_a$ and $t_b=t_a + \tau$. Let  $\tau'$ be a random variable where its distribution $F_{\tau'}(\cdot)$ is the probability of each difference between each sample of $\bar{J}_a$ and $\bar{J}_b$. The support of $\tau'$ is $\tau -N +1 \le \tau' \le \tau + N - 1$.
\end{definition}

\begin{theorem}[Auto-Correlation and Covariance]
\label{theorem:auto_corr_covariance}
    Let $\tau$ be the difference between two time steps of the processes $Y$. Let $\mathbb{E}_{\tau'}$ be an expectation according to the distribution $F_{\tau'}(\cdot)$. Then: 
    \begin{equation}\label{eq:R_Y_C_Y}
    \begin{split}
        R_Y(\tau) &= \mathbb{E}_{\tau'} \left[ R_Z(\tau') \right], \\
        C_Y(\tau) &= \mathbb{E}_{\tau'} \left[ C_Z(\tau')  \right].
    \end{split}
    \end{equation}
\end{theorem}
The proof for Theorem \ref{theorem:auto_corr_covariance} is in Section \ref{app:prove_theorem:auto_corr_covariance} in the supplementary material. We note two things. First, we note that we did not specify how the sampling is done from the RB and it is expressed by the random variable $\tau'$ from Definition \ref{definition:tau_tag}, i.e., Eq. \eqref{eq:R_Y_C_Y} is a general expression. Second, we note that we express the correlation using process $Z$ and not process $X$ directly, but process $Z$ auto-correlation and covariance can be computed directly in any practical case using the relation $Z_t=f(X_t)$. 

For the specific case  of "unordered sampling without replacement", we express the relation between the second moments of $Z$ and $Y$ explicitly through the distribution of $\tau'$.
\begin{lemma}[Time difference Distribution for uniform batches]
\label{lemma:difftime}
The random variable $\tau'$ distribution for "unordered sampling without replacement" is 
\begin{equation*}
    \begin{split}
        P(\tau') = 
        \begin{cases}
        \frac{N-|d|}{N^2} & \tau'=\tau + d, \\
        & d \in \{-N + 1, \ldots 0, \ldots N - 1\} \\
        0 & \tau' < \tau-N + 1 \text{ or } \tau' > \tau+N - 1
        \end{cases}.
    \end{split}
\end{equation*}
\end{lemma}
The proof for Lemma \ref{lemma:difftime} is in Section \ref{app:proof_lemma:difftime} in the supplementary material. In the following corollary we state the exact dependence in the case of random sampling of $\NBATCH$ samples from a RB with size $N$.
\begin{corollary}
\label{corr:autocorr_covariance_binom_sampling}
Consider process $Z$ where sampling is according to "unordered sampling without replacement". Then, the auto-correlation and covarinace of the process $Y$ are:
\[R_Y(\tau) = \frac{1}{N^2} \sum_{d=-N+1}^{N-1} (N-|d|) R_Z(d+\tau) \]
\[C_Y(\tau) = \frac{1}{N^2} \sum_{d=-N+1}^{N-1} (N-|d|) C_Z(d+\tau)\]
\end{corollary}

The proof for Corollary \ref{corr:autocorr_covariance_binom_sampling} is in Section \ref{app:proof_corr:autocorr_covariance_binom_sampling} in the supplementary material. We see that using a RB reduces the autocorrelation and covaraince of process $Z$ by factor of $N$. Interestingly,  this reduction is independent of $K$. This result proves the de-correlation effect of using RBs and provides an explicit expression for that.

\section{Replay Buffers in Reinforcement Learning}
\label{section:RB_in_RL}

In the previous section we analyzed properties of stochastic processes that go through a RB. In this section we analyze RBs in RL. Stabilizing learning in modern off-policy deep RL algorithms, such as Deep Q-Networks \cite{mnih2013playing} or DDPG \cite{lillicrap2015continuous}, is based on saving past observed transitions in a RB. Even though its use is extensive, the theoretical understanding of sampling batches mechanism from a RB is still quite limited. This is  our focus in this section. 

We begin with describing the setup that will serve us in this section. Then we connect between the random processes as defined in Section \ref{section:RBProperties} and common stochastic updates used in RL. We then describe an RB-based actor-critic algorithm that uses a batch of $K$ samples from the RB in each parameters update step. This type of algorithm serves as a basic example for popular usages of RBs in RL. We note that other versions of RB-based RL algorithms (such as deep RL algorithms, value-based algorithms, discounted settings of the value function) can be analyzed with the stochastic processes tools we provide in this work. Finally, we present a full convergence proof for the RB-based actor critic algorithm. 

Despite its popularity, to the best of our knowledge, there is only handful of proofs that consider RB in RL algorithm analysis (e.g., \citealp{di2021sim} or \citealp{lazic2021improved}).
Most of the convergence proofs for off-policy algorithms assume that a single sample is sampled from the RB. \citet{di2021sim} proved for the first time the convergence of an RB-based algorithm. However, their algorithm and technical tools were focused on the sim-to-real challenge with multiple MDP environments, and they focused only on a single sample batch from the RB instead of $K$ samples (which complicates the proof). Therefore, for completeness and focusing on the RB properties, we provide a proof for RB-based algorithms, with a single MDP environment and a batch of $K$ samples. Similarly to previous works, we consider here a setup with linear function approximation \cite{bertsekas1996neuro}.

\subsection{Setup for Markov Decision Process}
An environment in RL is modeled as a Markov Decision Process (MDP; \citealp{puterman1994markov}), where $\mathcal{S}$ and $\mathcal{A}$ are the state space and action space, respectively. We let $P(S'|S,A)$ denote the probability of transitioning from state $S \in \mathcal{S}$ to state $S'\in \mathcal{S}$ when applying action $A \in \mathcal{A}$.
We consider a probabilistic policy  $\pi_\theta(A|S)$, parameterized by $\theta \in \Theta \subset \mathbb{R}^d$ which expresses the probability of the agent to choose an action $A$ given that it is in state $S$. The MDP measure $P(S'|S, A)$ and the policy measure $\pi_\theta(A|S)$ induce together a Markov Chain (MC) measure $P_\theta(S'|S)$ ($P_\theta$ in matrix form). We let $\mu_{\theta}$ denote the stationary distribution induced by the policy $\pi_\theta$.  The reward function is denoted by $r(S,A)$.  

In this work we focus on the \emph{average reward} setting \footnote{The discount factor settings can be obtained in similar way to current setup.}. The goal of the agent is to find a policy that maximizes the average reward that the agent receives during its interaction with the environment. Under an ergodicity assumption, the average reward over time eventually converges to the expected reward under the stationary distribution \cite{bertsekas2005dynamic}:
\begin{equation}
\label{eq:actor_loss}
    \eta_{\theta} \triangleq \lim_{T \rightarrow \infty} \frac{\sum_{t=0}^T r(S_t, A_t)}{T} = \mathbb{E}_{S \sim \mu_\theta, A \sim \pi_\theta} [r(S,A)].
\end{equation}

The state-value function evaluates the overall expected accumulated rewards given a starting state $S$ and a policy $\pi_\theta$
\begin{equation}
\label{eq:V_definition}
    V^{\pi_\theta} (S)  \triangleq \mathbb{E} \left[ \left. \sum_{t=0}^{\infty} ( r(S_t, A_t) - \eta_{\theta} ) \right| S_0=S, \pi_\theta \right],
\end{equation}
where the actions follow the policy $A_t \sim \pi_\theta(\cdot|S_t)$ and the next state follows the transition probability $S_{t+1} \sim P(\cdot|S_t, A_t)$.

Let $\OO = \{S, A, S'\}$ be a transition from the environment. Let $\OO_t$ be a transition at time $t$. The temporal difference error $\delta (\OO)$ (TD; \citealp{bertsekas1996neuro}) is a random variable based on a single sampled transition from the $RB$,
\begin{equation}
\label{eq:TD_error}
    \delta(\OO) = r(S, A) - \eta + \phi(S')^\top w -  \phi(S)^\top w,
\end{equation}
where $\hat{V}^{\pi_\theta}_{w} (S) = \phi(S)^\top w $ is a linear approximation for $V^{\pi_\theta}(S)$,  $\phi(S) \in \mathbb{R}^d$ is a feature vector for state $S$ and $w \in \mathbb{R}^d$ is the critic parameter vector. We denote by $\Phi \in \mathbb{R}^{ |\mathcal{S}| \times d} $ the matrix of all state feature vectors.

\subsection{Replay Buffer as a Random Process in RL }
\label{section:connect_RB_RL_functions}

In Section \ref{section:RBProperties} we compared between properties of general random process $X$ going through a RB and yielding a process $Y$. In the RL context we have $X_t \triangleq \OO_t$, meaning our basic component is a single transition of state-action-next-state observed at time $t$. 
In addition, we defined $Z_t \triangleq f(X_t)$ process where $f(\cdot)$ is a general function. In RL, $f(\cdot)$ is commonly defined as the value function, the Q-function, the TD-error, the empirical average reward, the critic or actor gradients or any other function that computes a desirable update, based on an observed transition $\OO$. Common RL algorithms that use a single $f(\OO_t)$ computation in the parameters update step are commonly referred as \emph{on-policy} algorithms where they update their parameters based only on the last observed transition in the Markov chain. See Figure \ref{fig:RB_diagram} for a comparison between on-line updates and RB-based updates. Using the formulation of random processes we presented in Section   \ref{section:RBProperties}, we can characterize the on-line updates, based on a single last transition as follows: 
\begin{align*}
    Z_t^{\text{reward}} & = f_{\text{reward}}(\OO_t) = r(S_t, A_t) - \eta_t \\
    Z_t^{\text{critic}} & = f_{\text{critic}}(\OO_t) = \delta(\OO_t) \phi(S_t) \\
    Z_t^{\text{actor}} & = f_{\text{actor}}(\OO_t) = \delta(\OO_t) \nabla \log \pi_\theta(A_t|S_t)
\end{align*}

When using RB-based off-policy algorithms, the parameters updates are computed over an average of $\NBATCH$ functions which are based on $\NBATCH$ transitions that were sampled randomly from the last stored $N$ transitions. This exactly matches our definition of the process $Y$: $Y_t = \frac{1}{K} \sum_{j \in J_t} f(X_j) = \frac{1}{K} \sum_{j \in J_t} Z_j$. 
The following updates are typical in RB-based off-policy algorithms: 
\begin{equation}
\label{eq:Y_t_RL}
    \begin{split}
    Y_t^{\text{reward}} & = \frac{1}{K} \sum_{j \in J_t} Z_j^{\text{reward}} = \frac{1}{K} \sum_{j \in J_t} r(S_j, A_j) - \eta_t \\
    Y_t^{\text{critic}} & = \frac{1}{K} \sum_{j \in J_t} Z_j^{\text{critic}} = \frac{1}{K} \sum_{j \in J_t} \delta(\OO_j)  \phi(S_j) \\
    Y_t^{\text{actor}} & = \frac{1}{K} \sum_{j \in J_t} Z_j^{\text{actor}} = \frac{1}{K} \sum_{j \in J_t} \delta(\OO_j)  \nabla \log \pi_\theta(A_j|S_j)
    \end{split}
\end{equation}
In Algorithm \ref{alg:LACRB}, we present a linear actor critic algorithm based on RB samples where the algorithm updates the actor and critic using a random batch of transitions from the RB.  In Section \ref{sec:convergence} we show how the results from Section \ref{section:RBProperties} regarding a random process that is pushed into the RB, and the definitions of $X$ and $Y$ processes are helpful in proving the asymptotic convergence of this algorithm.

\subsection{Linear Actor Critic with RB Samples Algorithm}


\begin{algorithm}[t]
\caption{Linear Actor Critic with RB samples} 
\label{alg:LACRB}
\begin{algorithmic}[1]
\STATE Initialize Replay Buffer $\textrm{RB}$ with size $N$.
\STATE Initialize actor parameters $\theta_0$, critic parameters $w_0$ and average reward estimator $\eta_0$.
\STATE Learning steps $\{\alpha^{\eta}_t\}$, $\{\alpha^{w}_t\}$, $\{\alpha^{\theta}_t\}$.
\FOR{$t=0, \ldots$}
    \STATE Interact with MDP $M$ according to policy $\pi_{\theta_t}$ and add the transition $\{S_t, A_t, r(S_t, A_t), S_{t+1} \}$ to $\text{RB}_t$. 
    \STATE Sample $J_t$ - $\NBATCH$ random time indices form $RB_t$. Denote the corresponding transitions as $\{\OO_j\}_{j \in J_t}$. 
    \STATE $\delta(\OO_j) = r(S_j, A_j) - \eta_t + \phi(S'_j)^\top w_t  - \phi(S_j)^\top w_t$ \label{line:TD}
        \STATE Update average reward\\ $\eta_{t+1} = \eta_t + \alpha^{\eta}_t (\frac{1}{\NBATCH} \sum_{j \in J_t} r(S_j, A_j) - \eta_t) $ \label{line:average_reward}
        \STATE Update critic $w_{t+1} = w_{t} + \alpha^w_t \frac{1}{\NBATCH}  \sum_{j \in J_t} \delta(\OO_j) \phi(S_j)$ \label{line:critic}
        \STATE Update actor  $\theta_{t+1} = \Gamma \big( \theta_{t} - \alpha^{\theta}_t \frac{1}{\NBATCH}  \sum_{j \in J_t} \delta(\OO_j) \nabla_\theta \log \pi_\theta (A_j | S_j) \big)$ \label{line:actor}
\ENDFOR
\end{algorithmic}
\end{algorithm}

\begin{figure*}[t]
\centerline{\includegraphics[width=1.87\columnwidth]{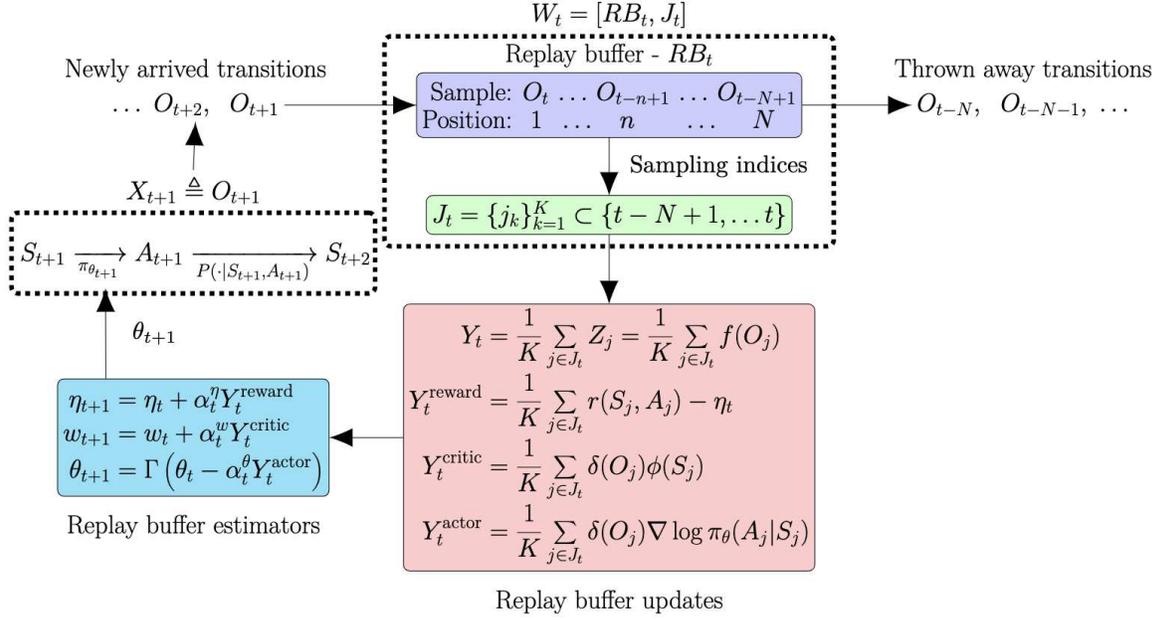}}
\caption{Replay buffer in reinforcement learning flow diagram: The random processes described in Figure \ref{fig:RB_diagram} are reflected in Algorithm \ref{alg:LACRB}. Here the random process that enters the RB is $\OO$ which is a tuple of $(S, A, S')$. The RB stores the last $N$ transitions $\{\OO_{t}, \ldots \OO_{t-N-1}\}$ in positions $(1, \ldots, N)$, respectively. As time proceeds and $t>N$, old transition are thrown away from the RB. At each time step $t$, a random subset of $K$ time steps is sampled from the RB and is denoted as $J_t$. $W$ is simply  $[RB, J]$. In Algorithm \ref{alg:LACRB} we have three different updates, $Y_t^{\text{reward}}$, $Y_t^{\text{critic}}$ and $Y_t^{\text{actor}}$, all are averages over functions of transitions sampled from the RB. Then the parameters are updated accordingly. Finally, the policy parameter $\theta_{t+1}$ is used to sample the action in transition $\OO_{t+1}$ that later enters to the RB.}
\label{fig:RB_diagram_RL}
\end{figure*}

The basic RB-based algorithm we analyze in this work is presented in Algorithm \ref{alg:LACRB}. We propose a two time scale linear actor critic optimization scheme (similarly to \citealp{konda2000actor}), which is an RB-based version of \citet{bhatnagar2008incremental} algorithm. Our algorithm is fully described by the random process $W_t=[RB_t, J_t]$ and by the algorithm updates $Y_t^{\text{reward}}$, $Y_t^{\text{critic}}$ and $Y_t^{\text{actor}}$ described in equation  \eqref{eq:Y_t_RL}. See Figure \ref{fig:RB_diagram_RL} for a visualized flow diagram of Algorithm \ref{alg:LACRB}.  

In Algorithm \ref{alg:LACRB} we consider an environment, modeled as an MDP $M$, and we maintain a replay buffer RB with capacity $N$. The agent collects transitions $\{S, A, r(S,A), S'\}$ from the environment and stores them in the RB. We train the agent in an off-policy manner. At each time step $t$, the agent samples $J_t$ -- a subset of $K$ random time indices from $\RB_t$ which defines the random transitions batch for optimizing the average reward, critic and actor parameters. Note that for the actor updates, we use a projection $\Gamma (\cdot)$ that projects any $\theta \in \mathbb{R}^d$ to a compact set $\Theta$ whenever $\theta \notin \Theta$.

\subsection{Expectations of Critic and Actor Updates in Algorithm \ref{alg:LACRB}}
\label{subsec:Algorithm_1_asymptotic_convergence}
We divide the convergence analysis of Algorithm \ref{alg:LACRB} into two parts. The first part, presented in this section, is unique to our paper - we describe in Lemmas \ref{lemma:expected_delta} and \ref{lemma:expected_delta_grad_log} a closed form of the expectations of the actor and critic updates, based on a random batch of $K$ transitions from the RB. In the second part, presented in Section \ref{sec:convergence}, we use Stochastic Approximation tools for proving the algorithm updates convergence, based on the results from Lemmas \ref{lemma:expected_delta} and \ref{lemma:expected_delta_grad_log}. We note that Section \ref{sec:convergence} follows the steps of the convergence proofs presented by \citet{di2021sim} and \citet{bhatnagar2009natural}.

For time $t - n + 1$ where $1 \le n \le N$, we define the induced MC with a corresponding policy parameter $\theta_{t - n +1}$. For this parameter, we denote the corresponding state distribution vector $\rho_{t - n + 1}$ and a transition matrix $P_{t - n + 1}$ (both induced by the policy $\pi_{\theta_{t - n + 1}}$). Finally, we define the following diagonal matrix $D_{t - n + 1}\triangleq \textrm{diag}(\rho_{t - n +1})$ and the reward vector $r_{t - n +1}$ with elements $r_{t - n +1}(S) = \sum_{A} \pi_{\theta_{t - n +1}} (A|S) r(S,A)$. Based on these definitions we define
\begin{equation}
\label{eq:def_A_b_of_TD}
    \begin{split}
        C_t 
        & \triangleq     
        \frac{1}{N} \sum_{n=1}^N 
        D_{t - n +1} \left( P_{t - n +1} - I \right) \\ 
        b_t 
         & \triangleq     
        \frac{1}{N} \sum_{n=1}^N 
        D_{t- n + 1} \left( r_{t - n + 1} - \eta_{\theta} e \right).\\
    \end{split}
\end{equation}
where $I$ is the identity matrix and $e$ is a vector of ones. 
Let $D_{\theta} \triangleq \textrm{diag}(\mu_{\theta})$ and define
\begin{equation}
\label{eq:def_A_b_of_TD_infty}
\begin{split}
    C_\theta &\triangleq D_{\theta} \left( P_{\theta} - I \right), \quad
    b_\theta \triangleq  D_{\theta} \left( r_{\theta} - \eta_{\theta} e \right).
\end{split}
\end{equation}
In our RB setting, since we have at time $t$ a RB with the last $N$ samples, $C_t$ and $b_t$ in Equation \eqref{eq:def_A_b_of_TD} represent a superposition of all related elements of these samples. For proving the convergence of the critic, we assume the policy is fixed. Then, when $t \rightarrow \infty$,  $\rho_{t - n + 1} \rightarrow \mu_{\theta}$ for all index $n$. This means that the induced MC is one for all the samples in the RB, so the sum over $N$ disappears for $C_\theta$ and $b_\theta$.

The following two lemmas compute the expectation of the critic and actor updates when using a random batch of $K$ samples. The expectations are over all possible random batches sampled from the RB. Recall that $\bar{J}_t \subset \{1, \ldots, N\}$ and $\mathbb{C}_{N,\NBATCH}$ is the set of all possible subsets $\bar{J}$ for specific $N$ and $\NBATCH$. These lemmas are the main results for proving convergence of RB-based RL algorithms. 
\begin{lemma}
\label{lemma:expected_delta}
Assume we have a RB with $N$ transitions and we sample random $K$ transitions from the RB. Then: 
\begin{equation*}
\begin{split}
    & \mathbb{E}_{\bar{J}_t \sim \mathbb{C}_{N, K}, \{\OO_{t-n+1}\}_{n \in \bar{J}_t} \sim\text{RB}_t} \\
    & \left[  \frac{1}{\NBATCH} \sum_{n \in \bar{J}_t} \delta(\OO_{t - n + 1}) \phi(S_{t-n+1}) \right]  = \Phi^\top C_\theta \Phi w +  \Phi^\top b_\theta,
\end{split}
\end{equation*}
where $C_\theta$ and $b_\theta$ are defined in \eqref{eq:def_A_b_of_TD_infty}. 
\end{lemma}

\begin{lemma}
\label{lemma:expected_delta_grad_log}
Assume we have a RB with $N$ transitions and we sample random $K$ transitions from the RB. Then:
\begin{equation*}
    \begin{split}
        & \mathbb{E}_{\bar{J}_t \sim \mathbb{C}_{N, K}, \{\OO_{t-n+1}\}_{n \in \bar{J}_t} \sim\text{RB}_t} \\
        & \left[  \frac{1}{\NBATCH} \sum_{n \in \bar{J}_t} \delta^{\pi_\theta}(\OO_{t - n + 1}) \nabla_\theta \log \pi_{\theta} (A_{t - n + 1}|S_{t - n + 1}) \right]   \\
        & = \nabla_\theta \eta_\theta - \sum_{S} \mu_{\theta}  (S) \Big(  \phi(S)^\top \nabla_\theta w^{\pi_\theta} -  \nabla_\theta \bar{V}^{\pi_{\theta}}(S) \Big),
    \end{split}
\end{equation*}
where $\bar{V}^{\pi_{\theta}}(S) = \sum_{A \in \mathcal{A}} \pi_{\theta} (A|S) \big( r(S,A) - \eta_{\theta}$ $ + \sum_{S' \in \mathcal{S}} P (S'|S, A) \phi (S')^\top w^{\pi_\theta} \big)$.
\end{lemma}

The proofs for Lemmas \ref{lemma:expected_delta} and \ref{lemma:expected_delta_grad_log} are in sections \ref{app:proof_lemma_expected_delta} and \ref{app:proof of actor lemma}, respectively, in the supplementary material.

\subsection{Asymptotic Convergence of Algorithm \ref{alg:LACRB}}
\label{sec:convergence}
We are now ready to present the convergence theorems for the critic and actor in Algorithm \ref{alg:LACRB}. In the proof of our theorems we use tools from Stochastic Approximation (SA) \cite{kushner2003stochastic, borkar2009stochastic, bertsekas1996neuro}, a standard tool in the literature for analyzing iterations of processes such as two time scale Actor-Critic in the context of RL. 



We showed in Lemma \ref{lemma:RB_is_markovian} that the process $W_t=[RB_t, J_t]$ of sampling $K$ random transitions from the RB is a Markov process. In addition, we showed in Lemma \ref{lemma: ergodicity}  that if the original Markov chain is irreducible and aperiodic, then also the RB Markov process is irreducible and aperiodic. This property is required for the existence of unique stationary distribution and for proving the convergence of the iterations in Algorithm \ref{alg:LACRB} using SA tools. We note that proving convergence for a general function approximation is hard. We choose to demonstrate the convergence for a linear function approximation (LFA; \citealp{bertsekas1996neuro}), in order to keep the convergence proof as simple as possible while focusing in the proof on the RB and random batches aspects of the algorithm. 


We present several assumptions that are necessary for proving the convergence of Algorithm \ref{alg:LACRB}. Assumption \ref{ass:independent_features} is needed for the uniqueness of the convergence point of the critic. In addition, we choose a state $S^*$ to be of value $0$, i.e., $V^{\pi_\theta} (S^*)=0$ (due to Assumption \ref{ass:aperiodic}, $S^*$ can be any of $S\in \mathcal{S}$). Assumption \ref{ass:step_size} is required in order to get a \emph{with probability 1} using the SA convergence. In our actor-critic setup we need two time-scales convergence, thus, in this assumption the critic is a ‘faster’ recursion than the actor.

\begin{assumption}
\label{assumption:theta_is_compact}
1. The set $\Theta$ is compact.
2. The reward $|r(\cdot,\cdot)| \le 1$ for all $S\in \mathcal{S}, A\in \mathcal{A}$.
\end{assumption}

\begin{assumption}
\label{ass:aperiodic}
For any policy $\pi_\theta$, the induced Markov chain of the MDP process $\{S_t\}_{t\ge 0}$ is
irreducible and aperiodic.
\end{assumption}

\begin{assumption}
\label{ass:continous_policy}
For any state–action pair $(S,A)$, $\pi_\theta(A|S)$ is continuously differentiable in the parameter $\theta$.
\end{assumption}

\begin{assumption}
\label{ass:independent_features}
1. The matrix $\Phi$ has full rank. 
2. The functions $\phi(S)$ are Liphschitz in $S$ and bounded. 
3. For every $w \in \mathbb{R}^d$, $\Phi w \neq e$ where $e$ is a vector of ones. 
\end{assumption}

\begin{assumption}
\label{ass:step_size}
The step-sizes $\{ \alpha^{\eta}_t\}$, $\{ \alpha^{w}_t\}$, $\{ \alpha^{\theta}_t\}$, $t \ge 0$ satisfy $\sum_t^\infty \alpha^{\eta}_t = \sum_t^\infty \alpha^{w}_t = \sum_t^\infty \alpha^{\theta}_t = \infty$, $\sum_t^\infty (\alpha^{\eta}_t)^2,\quad  \sum_t^\infty (\alpha^{w}_t)^2, \quad \sum_t^\infty (\alpha^{\theta}_t)^2 < \infty$ and $\alpha^{\theta}_t = o (\alpha^{w}_t)$.
\end{assumption}
Now we are ready to prove the following theorems, regarding Algorithm \ref{alg:LACRB}. We note that Theorem \ref{theorem:critic_conv} and \ref{theorem:actor_conv} state the critic and actor convergence.
\begin{theorem}
\label{theorem:critic_conv} (Convergence of the Critic to a fixed point)\\
Under Assumptions \ref{assumption:theta_is_compact}-\ref{ass:step_size}, for any given $\pi$ and $\{\eta_t\}, \{w_t\}$ as in the updates in Algorithm \ref{alg:LACRB}, we have $\eta_t \rightarrow \eta_\theta $ and $w_t \rightarrow w^\pi $ with probability 1, where $w^\pi$ is obtained as a unique solution to 
$\Phi^\top C_\theta \Phi w + \Phi^\top b_\theta = 0$. 
\end{theorem}
The proof for Theorem \ref{theorem:critic_conv} is in Section \ref{appendix:critic_conv} in the supplementary material. It follows the proof for Lemma 5 in \citet{bhatnagar2009natural}.
For establishing the convergence of the actor updates, we define additional terms. Let $\mathcal{Z}$ denote the set of asymptotically stable equilibria of the ODE $\dot \theta = \hat \Gamma (- \nabla_\theta \eta_\theta)$ and let $\mathcal{Z}^\epsilon$ be the $\epsilon$-neighborhood of $\mathcal{Z}$. We define $\xi^{\pi_\theta} = \sum_{S} \mu_{\theta}  (S) \Big(  \phi(S)^\top \nabla_\theta w^{\pi_\theta} -  \nabla_\theta \bar{V}^{\pi_{\theta}}(S) \Big)$.
\begin{theorem}
\label{theorem:actor_conv} (Convergence of the actor)\\
Under Assumptions \ref{assumption:theta_is_compact}-\ref{ass:step_size}, given $\epsilon > 0$, $\exists \delta > 0$ such that for $\theta_t$, $t \ge 0$ obtained using Algorithm \ref{alg:LACRB}, if $\sup_{\theta_t} \| \xi^{\pi_{\theta_t}}\| < \delta $, then $\theta_t \rightarrow \mathcal{Z}^\epsilon$ as $t \rightarrow \infty$ with probability one. 
\end{theorem}
The proof for Theorem \ref{theorem:actor_conv} is in Section \ref{appendix:actor_conv} in the supplementary material. It follows the proof for Theorem 2 in \citet{bhatnagar2009natural}.

\section{Related Work}\label{section:related_work}

\textbf{Actor critic algorithms analysis:} The convergence analysis of our proposed RB-based actor critic algorithm is based on the Stochastic Approximation method \cite{kushner2012stochastic}. \citet{konda2000actor} proposed the actor-critic algorithm, and established the asymptotic convergence for the two time-scale actor-critic, with TD($\lambda$) learning-based critic. \citet{bhatnagar2009natural} proved the convergence result for the original actor-critic and natural actor-critic methods. \citet{dicastro2010convergent} proposed a single time-scale actor-critic algorithm and proved its convergence. Works on finite sample analysis for actor critic algorithms \cite{wu2020finite, zou2019finite,dalal2018finite} analyze the case of last transition update and do not analyze the RB aspects in these algorithms. 

Recently, \citet{di2021sim} proved for the first time the convergence of an RB-based actor critic algorithm. However, their algorithm and technical tools were focused on the sim-to-real challenge with multiple MDP environments, and they focused only on a single sample batch from the RB instead of $K$ samples. 
We provide a proof for RB-based algorithms, with a single MDP environment and a batch of $K$ samples. 

\textbf{Replay Buffer analysis:} Experience replay \cite{lin1993reinforcement} is a central concept for achieving good performance in deep reinforcement learning. Deep RB-based algorithms such as deep Q-learning (DQN, \citealp{mnih2013playing}), deep deterministic policy gradient (DDPG; \citealp{lillicrap2015continuous}), actor critic with experience replay (ACER; \citealp{wang2016sample}), Twin Delayed Deep Deterministic policy gradient (TD3, \citealp{fujimoto2018addressing}), Soft Actor Critic (SAC, \citealp{haarnoja2018soft}) and many others use RBs to improve performance and data efficiency. 

We focus mainly on works that provide some RB properties analysis. \citet{zhang2017deeper} and \citet{liu2018effects} study the effect of replay buffer size on the agent performance . \citet{fedus2020revisiting} investigated through simulated experiments how the learning process is affected by the ratio of learning updates to experience collected.  Other works focus on methods to prioritize samples in the RB and provide experimental results to emphasis performance improvement when using prioritized sampling from RB \cite{schaul2015prioritized,pan2018organizing,zha2019experience,horgan2018distributed,lahire2021large}. We, on the other hand, focus on the theoretical aspects of RB properties and provide convergence results for RB-based algorithms. 
\citet{lazic2021improved} proposed a RB version for a regularized policy iteration algorithm. They provide an additional motivation for using RBs, in addition to the advantage of reduced temporal correlations: They claim that using RB in online learning in MDPs can approximate well the average of past value functions. Their analysis also suggests a new objective for sub-sampling or priority-sampling transitions in the RB, which differs priority-sampling objectives of previous work \cite{schaul2015prioritized}.

Regarding RB analysis in Deep RL algorithms, \citet{fan2020theoretical} performed a finite sample analysis on DQN algorithm \cite{mnih2013playing}. In their analysis, they simplified the technique of RB with an independence assumption and they replaced the distribution over random batches with a fixed distribution. These assumptions essentially reduce DQN to the neural fitted Q-iteration (FQI) algorithm \cite{riedmiller2005neural}. In our work we focus on asymptotic convergence and analyze explicitly the distribution of random batches from the RB.

\section{Conclusions} \label{section:conclusions}
In this work we analyzed RB and showed some basic random processes properties of it as ergodicity, stationarity, Markovity, correlation, and covariance. The latter two are of most interest since they can explain the success of modern RL algorithm based on RB. 
In addition, we developed theoretical tools of stochastic process analysis for replay buffers. We provided an example of how to use these tools to analyze the convergence of an RB-based actor critic algorithm. Similarly, other common RB-based algorithms in reinforcement learning such as DQN \cite{mnih2013playing}, DDPG \cite{lillicrap2015continuous}, TD3 \cite{fujimoto2018addressing}, SAC \cite{haarnoja2018soft} and many others can be analyzed now, using the tools we developed in this work. 

As a future research, we propose two directions that are of great interest and complete the analysis we provided in this work:
\begin{enumerate}
    \item \textbf{Spectrum analysis of the learning processes.} Since we adopted an approach of "Signals and Systems" with random signals (\citealp{oppenheim1997signals,porat2008digital}), one can use spectrum analysis in order to discover instabilities or cycles in the learning process.
    \item \textbf{More complex RBs.} There is a large experimental body of work that tries to propose different schemes of RBs. Some of them apply different independent on RL quantities sampling techniques while other  apply dependent on RL quantities schemes (e.g., prioritized RB depends on the TD signal; \citealp{schaul2015prioritized}). In this work we paved the first steps to apply analysis on such schemes (both dependent and independent).
\end{enumerate}

\section{Acknowledgements}
This work was partially supported by the Israel Science Foundation under contract 2199/20. 

\bibliography{main}
\bibliographystyle{icml2022} 

\newpage
\onecolumn

\appendix
\numberwithin{equation}{section}
\section{Proofs for Lemmas in Section \ref{section:RBProperties}}
\label{appendix:proof_RBProperties}

\subsection{Proof of Lemma \ref{lemma:Yt_is_stationary}}
\label{appendix:proof_stationarity}

\begin{proof}
\textbf{Stationarity of $RB_t$:}
Recall that stationarity (in the strong sense) means that for $m=1,2,\dots$, there are times $(t_1, t_2, \ldots, t_m)$ such that for all $\tau \in \mathbb{Z}$
\begin{equation*}
\label{app_eq:stationary}
    F_X(X_{{t_1+\tau}},\ldots ,X_{{t_m+\tau}}) =F_X(X_{t_1},\ldots ,X_{t_m}),
\end{equation*}
where $F_X(X_{t_1 },\ldots ,X_{t_m })$ is the cumulative distribution. Then,
\begin{equation*}
\begin{split}
    F_{\RB}(\RB_{{t_1+\tau}},\ldots ,\RB_{{t_m+\tau}}) 
    \stackrel{(1)}{=} &F_{X}(X_{{t_1+\tau-N+1}},\ldots ,X_{{t_1+\tau}},  \\
      &X_{{t_2+\tau-N+1}},\ldots ,X_{{t_2+\tau}},  \\
      &\ldots,\\
      &X_{{t_m+\tau-N+1}},\ldots ,X_{{t_m+\tau}}),  \\
    \stackrel{(2)}{=} &F_{X}(X_{{t_1-N+1}},\ldots ,X_{{t_1}},  \\
      &X_{{t_2-N+1}},\ldots ,X_{{t_2}},  \\
      &\ldots,\\
      &X_{{t_m-N+1}},\ldots ,X_{{t_m}})  \\
    \stackrel{(3)}{=} &F_{\RB}(\RB_{{t_1}},\ldots ,\RB_{{t_m}}),  \\
\end{split}
\end{equation*}
where we use the RB definition in (1), stationarity of $X$ in (2), and expressing RB based on $X$ in (3).

\noindent \textbf{Stationarity of $Y_t$:} Similarly, for $m=1,2,\dots$, there are times $(t_1, t_2, \ldots, t_m)$ such that for all $\tau \in \mathbb{Z}$
\begin{equation*}
\begin{split}
    F_{Y}(Y_{{t_1+\tau}},\ldots ,Y_{{t_m+\tau}}) 
    \stackrel{(1)}{=} 
    &F_{X}\left(\frac{1}{K}\sum_{j \in \bar{J}_{t_1+\tau}}f(X_j),\ldots,
    \frac{1}{K}\sum_{j \in \bar{J}_{t_m+\tau}}f(X_j)\right)\\
    \stackrel{(2)}{=} 
    &\sum_{\bar{J}_{t_1+\tau},\ldots, \bar{J}_{t_m+\tau}}
    F_{\bar{J}_{t_1+\tau},\ldots, \bar{J}_{t_m+\tau}}(j_1,\ldots, j_m) \times\\
    &F_{X}\left(\left.\frac{1}{K}\sum_{j \in \bar{J}_{t_1+\tau}}f(X_j),\ldots,
    \frac{1}{K}\sum_{j \in \bar{J}_{t_m+\tau}}f(X_j)\right|j_1,\ldots, j_m \right)\\
    \stackrel{(3)}{=} 
    &\sum_{\bar{J}_{t_1},\ldots, \bar{J}_{t_m}}
    F_{\bar{J}_{t_1},\ldots, \bar{J}_{t_m}}(j_1,\ldots, j_m) \times\\
    &F_{X}\left(\left.\frac{1}{K}\sum_{j \in \bar{J}_{t_1}}f(X_j),\ldots,
    \frac{1}{K}\sum_{j \in \bar{J}_{t_m}}f(X_j)\right|j_1,\ldots, j_m \right)\\
    \stackrel{(4)}{=} 
    &F_{X}\left(\frac{1}{K}\sum_{j \in \bar{J}_{t_1}}f(X_j),\ldots,
    \frac{1}{K}\sum_{j \in \bar{J}_{t_m}}f(X_j)\right)\\
    \stackrel{(5)}{=} 
    &F_{Y}(Y_{{t_1}},\ldots ,Y_{{t_m}}),
    \end{split}
\end{equation*}
where in (1) we use the process $Y$ definition, in (2) we use iterated expectation, in (3) we use both the stationarity of $X$ and $\bar{J}$, in (4) we use again iterated expectation, and in (5) we use $Y$ definition.

\end{proof}

\subsection{Proof of Lemma \ref{lemma:RB_is_markovian}}
\label{appendix:proof_RB_is_markovian}
\begin{proof}
We first note that we use some abuse of notation when referring sometimes to random variables from processes $X$, $RB$, $W$ and $J$ the same as their realizations. However, to avoid an overhead of the proof, we keep the notations simple and short. 

Proving Markovity requires that
\begin{align}
    \label{eq:RB_definition_of_markov}
    P(\RB_{t+1}|\RB_{t}, \RB_{t-1}, \ldots, \RB_0) & = P(\RB_{t+1}|\RB_{t}).\\
    \label{eq:W_definition_of_markov}
    P(W_{t+1}|W_{t}, W_{t-1}, \ldots, W_0) & = P(W_{t+1}|W_{t}).
\end{align}
We start with proving the Markovity of $\RB_t$. Let us denote $\X_{n_1}^{n_2} (t) \triangleq \{\X_{t - n_2 + 1}, \ldots, \X_{t - n_1 + 1} | \RB_t\}$ as the set of realizations of the random variables from process $\X$ stored in the RB at time $t$ in positions $n_1$ to $n_2$. 
Note that when a new transition is pushed to the RB into position $n=1$, the oldest transition in position $n=N$ is thrown away, and all the transitions in the RB move one index forward. 
We present here some observations regarding the RB that will help us through the proof:
\begin{align}
    \label{eq:RB_definition} RB_t & = X_{1}^N (t) =  \{\X_{t-N+1}, \ldots, \X_{t-n+1} \ldots,  \X_t\} && \text{ (RB definition).} \\
    \label{eq:X_union} \X_{1}^{N}(t+1) & = \{\X_{t+1}\} \cup \X_{1}^{N-1}(t) \\
    \label{eq:X_subset} X_{1}^{N-1}(t) & \subset  \X_{1}^{N}(t) \\
    \label{eq:X_include} X_t & \in X_1^N(t) \\
    \label{eq:X_Markovian} P(X_{t+1} | X_t, \ldots X_0) & = P(X_{t+1} | X_t) && \text{ (Since } X_t \text{ is assumed to be Markovian). } \\
    \label{eq:joint_cond} P(a, b|c_1, c_2, \ldots) &= P(a | b, c_1, c_2, \ldots) \cdot P(b|c_1, c_2, \ldots) && \text{ (Expressing joint probability } \\
    \nonumber & && \text{ as a conditional probabilities product). } \\
    \label{eq:cond_independent} P(a | b) & = p(a) && \text{ (If a and b are independent) }. 
\end{align}
Computing the l.h.s. of equation  \eqref{eq:RB_definition_of_markov} yields
\begin{equation*}
    \begin{split}
        P\left(\RB_{t+1} \middle| \RB_{t}, \ldots, \RB_{0}\right) & \stackrel{\eqref{eq:RB_definition}}{=}  
        P \left(\X_{1}^{N}(t+1) \middle| \X_{1}^{N}(t),  \ldots, \X_{1}^{N}(0) \right)\\
        & \stackrel{\eqref{eq:X_union}}{=}   P \left(\X_{t+1}, \X_{1}^{N-1}(t) \middle| \X_{1}^{N}(t),  \ldots, \X_{1}^{N}(0) \right)\\
        & \stackrel{\eqref{eq:joint_cond}}{=}   P \left(\X_{t+1} \middle| \X_{1}^{N-1}(t),  \X_{1}^{N}(t),  \ldots, \X_{1}^{N}(0) \right) \cdot P \left(\X_{1}^{N-1}(t) \middle| \X_{1}^{N}(t),  \ldots, \X_{1}^{N}(0) \right)\\ 
        & \stackrel{\eqref{eq:X_subset},  \eqref{eq:X_include}, \eqref{eq:X_Markovian}}{=}   P \left(\X_{t+1} \middle| \X_t \right) 
    \end{split}
\end{equation*}        
Similarly, computing the r.h.s of \eqref{eq:RB_definition_of_markov} yields
\begin{equation*}
    \begin{split}
        P \left(RB_{t+1} \middle| RB_{t} \right)
        &\stackrel{\eqref{eq:RB_definition}}{=}   P \left(X_{1}^{N} (t+1)  \middle| X_{1}^{N} (t) \right) \\
        & \stackrel{\eqref{eq:X_union}}{=}   
        P \left(X_{t+1}, X_{1}^{N-1} (t)  \middle| X_{1}^{N} (t) \right) \\
        & \stackrel{\eqref{eq:joint_cond}}{=}   P \left(X_{t+1}   \middle| X_{1}^{N-1}(t), X_{1}^{N} (t) \right) \cdot  P \left(X_{1}^{N-1} (t)  \middle| X_{1}^{N} (t) \right) \\
        & \stackrel{\eqref{eq:X_subset},  \eqref{eq:X_include}, \eqref{eq:X_Markovian}}{=}   P \left(X_{t+1}   \middle| \X_t \right) 
    \end{split}
\end{equation*}
Both sides of \eqref{eq:RB_definition_of_markov} are equal and therefore $RB_t$ is Markovian. In addition we have that for $t \ge N$:
\[ P \left(RB_{t+1} \middle| RB_{t} \right) = \begin{cases}
P \left(X_{t+1}   \middle| \X_t \right) & \text{ if } X_t \in X_1^1(t) \text{ and } RB_{t+1} = \{X_{t+1}\} \cup X_1^{N-1}(t) \\
0  & \text{ otherwise }
\end{cases}.
 \]
Next, we prove the Markovity of $W_t$. Recall that $W_t$ is defined as:
\begin{equation}
\label{eq:W_definition}
    W_t = [RB_t, J_t]
\end{equation}
where $J_t \subset \{t-N+1, \ldots, t\}$ is a random subset of $K$ time indices. By their definition, $RB_t$ and $J_t$ are independent for all $t$. Computing the l.h.s. of equation  \eqref{eq:W_definition_of_markov} yields
\begin{equation*}
    \begin{split}
        P\left(W_{t+1} \middle| W_{t}, \ldots, W_{0}\right) & \stackrel{\eqref{eq:W_definition}}{=} P\left(RB_{t+1}, J_{t+1} \middle| RB_{t}, J_{t}, \ldots, RB_{0}, J_0 \right) \\
        & \stackrel{\eqref{eq:joint_cond}}{=} P \left(RB_{t+1} \middle| J_{t+1}, RB_{t}, J_{t}, \ldots, RB_{0}, J_0 \right) \cdot  P \left( J_{t+1} \middle| RB_{t}, J_{t}, \ldots, RB_{0}, J_0 \right) \\
        & \stackrel{\eqref{eq:RB_definition_of_markov}, \eqref{eq:cond_independent}}{=} P \left(RB_{t+1} \middle| RB_{t} \right) \cdot  P \left( J_{t+1} \right) \\
        & \stackrel{\eqref{eq:cond_independent}}{=} P \left(RB_{t+1}, J_{t+1}  \middle| RB_{t}, J_t \right) \\
        & \stackrel{\eqref{eq:W_definition}}{=} P \left( W_{t+1} \middle| W_t \right) 
    \end{split}
\end{equation*}   
We have the required result in \eqref{eq:W_definition_of_markov}, therefore $W_t$ is Markovian. In addition, If $J_{t+1}$ is sampled according to "unordered sampling without replacement" (defined in Section \ref{sec:smampling_method}), then for $t \ge N$:
\[P \left( W_{t+1} \middle| W_t \right) = P \left(RB_{t+1} \middle| RB_{t} \right) \cdot  P \left( J_{t+1} \right) = \begin{cases}
\frac{1}{\binom{N}{K}} P \left(X_{t+1}   \middle| \X_t \right) & \text{ if } X_t \in X_1^1(t) \text{ and } RB_{t+1} = \{X_{t+1}\} \cup X_1^{N-1}(t) \\
& \forall J_{t+1} \in \mathbb{C}_{N, K}, \\ 
0  & \text{ otherwise. }
\end{cases}.
\]
\end{proof}

\subsection{Proof of Lemma \ref{lemma: ergodicity}}
\label{appendix:proof_ergodicity}

\begin{proof}
We prove by contradiction. Let us assume that the process $\RB$ is neither aperiodic nor irreducible. If it is periodic, then one of the indices in the RB is periodic. Without loss of generality, let us assume that this is the $l$ delayed time-steps index. But since in this index we have a periodic process, i.e., it is the process $X$ delayed in $l$ steps, it contradicts the assumption that $X$ is aperiodic. We prove irreducibility in a similar way. 

Since the process $Y$ is a deterministic function of the process $\RB$, it must be aperiodic and irreducible as well, otherwise it will contradict the aperiodicity and irreducibility of the process $\RB$. 
Finally, since $f(\cdot)$ is a deterministic function, and since for each $t$, $Y_t$ is an image of an ergodic process $X_t$, i.e., each $x \in X_t$ is visited infinitely often, and as a results each point $y \in \text{supp}(Y_t)$ of the image of $f(\cdot)$ is visited infinitely often, otherwise, it contradicts the deterministic nature of $f(\cdot)$ or the ergodicity of $X$.
\end{proof}

\newpage
\section{Auto Correlation and Covariance proofs}
\label{app_sec:AutoCorr}

\subsection{Proof of Theorem  \ref{theorem:auto_corr_covariance}}
\label{app:prove_theorem:auto_corr_covariance}
\begin{proof}
Let $\bar{J}_t \subset \{1, \ldots N\}$ and $\bar{J}_{t+\tau} \subset \{1, \ldots N\}$ be subsets of $K$ indices each. We begin with calculating the auto-correlation of process $Z_t$.
\begin{equation*}
    \begin{split}
        R_Y(\tau) & = \mathbb{E}[Y_t Y_{t+\tau}] \\
            & \stackrel{(1)}{=} \mathbb{E} \left[ \frac{1}{\NBATCH} \sum_{n \in \bar{J}_t} Z_{t - n +1} \cdot \frac{1}{\NBATCH} \sum_{m \in \bar{J}_{t+\tau}} Z_{t + \tau - m +1} \right] \\
            & \stackrel{(2)}{=} \mathbb{E} \left[ \frac{1}{\NBATCH} \sum_{n \in \bar{J}_t} f(\X_{t - n +1}) \cdot \frac{1}{\NBATCH} \sum_{m \in \bar{J}_{t+\tau}} f(\X_{t + \tau - m +1}) \right] \\
            & \stackrel{(3)}{=} \mathbb{E}_{\bar{J}_t, \bar{J}_{t+\tau} \sim \mathbb{C}_{N,K}, \{\X_{t - n +1}\}_{n \in \bar{J}_t} \sim RB_t,  \{\X_{t + \tau - m +1}\}_{n \in \bar{J}_{t+ \tau}} \sim RB_{t + \tau} } \left[ \frac{1}{\NBATCH} \sum_{n \in \bar{J}_t} f(\X_{t - n +1}) \cdot \frac{1}{\NBATCH} \sum_{m \in \bar{J}_{t+\tau}}f(\X_{t + \tau - m +1}) \right] \\
            & \stackrel{(4)}{=} \mathbb{E}_{\bar{J}_t, \bar{J}_{t+\tau} \sim \mathbb{C}_{N,K}} \Bigg[  \mathbb{E}_{\{\X_{t - n +1}\}_{n \in \bar{J}_t} \sim RB_t,  \{\X_{t + \tau - m +1}\}_{n \in \bar{J}_{t+ \tau}} \sim RB_{t + \tau} } \\
            & \bigg[ \frac{1}{\NBATCH} \sum_{n \in \bar{J}_t} f(\X_{t - n +1}) \cdot \frac{1}{\NBATCH} \sum_{m \in \bar{J}_{t+\tau}} f(\X_{t + \tau - m +1})  \Bigg| \bar{J}_t, \bar{J}_{t+\tau} \bigg] \Bigg] \\
            & \stackrel{(5)}{=} \mathbb{E}_{\bar{J}_t, \bar{J}_{t+\tau} \sim \mathbb{C}_{N,K}} \Bigg[  \mathbb{E}_{\{\X_{t - n +1}\}_{n \in \bar{J}_t} \sim RB_t,  \{\X_{t + \tau - m +1}\}_{n \in \bar{J}_{t+ \tau}} \sim RB_{t + \tau} } \\
            & \bigg[ \mathbb{E}_{n \sim \bar{J}_t, m \sim \bar{J}_{t + \tau}} \left[ f(\X_{t - n +1}) \cdot f(\X_{t + \tau - m +1})  \right]  \Bigg| \bar{J}_t, \bar{J}_{t+\tau} \bigg] \Bigg] \\
            & \stackrel{(6)}{=} \mathbb{E}_{\bar{J}_t, \bar{J}_{t+\tau} \sim \mathbb{C}_{N,K}} \left[  \mathbb{E}_{n \sim \bar{J}_t, m \sim \bar{J}_{t + \tau}}  \left[    \mathbb{E}_{\X_{t - n +1},  \X_{t + \tau - m +1} }   \left[ f(\X_{t - n +1}) \cdot f(\X_{t + \tau - m +1})  \right]  \middle| \bar{J}_t, \bar{J}_{t+\tau} \right] \right] \\
            & \stackrel{(7)}{=} \mathbb{E}_{\bar{J}_t, \bar{J}_{t+\tau} \sim \mathbb{C}_{N,K}} \left[  \mathbb{E}_{n \sim \bar{J}_t, m \sim \bar{J}_{t + \tau}}  \left[    \mathbb{E} \left[ Z_{t - n +1} Z_{t + \tau - m +1}  \right]  \middle| \bar{J}_t, \bar{J}_{t+\tau} \right] \right] \\
            & \stackrel{(8)}{=} \mathbb{E}_{\bar{J}_t, \bar{J}_{t+\tau} \sim \mathbb{C}_{N,K}} \left[  \mathbb{E}_{n \sim \bar{J}_t, m \sim \bar{J}_{t + \tau}}  \left[    R_Z(\tau + n - m) \middle| \bar{J}_t, \bar{J}_{t+\tau} \right] \right] \\
            & \stackrel{(9)}{=}  \mathbb{E}_{\tau' \sim \tilde{J}_\tau }  \left[    R_Z(\tau') \right] 
    \end{split}
\end{equation*}
where in (1) we used the definition of Y using the indices subsets $\bar{J}_t$ and $\bar{J}_{t+\tau}$. In (2) used the definition of $Z$ and in (3) we wrote the expectation explicitly. In (4) we used the conditional expectation and in (5) we wrote $\frac{1}{K} \sum_{n \in \bar{J}_t} f(\cdot)$ and $\frac{1}{K} \sum_{m \in \bar{J}_{t+\tau}} f(\cdot)$ as an expectations since given the subsets  $\bar{J}_t$ and $\bar{J}_{t+\tau}$, the probability of sampling index $n$ or $m$ from the RB is uniform and equals $\frac{1}{K}$. In (6) we are left with the marginal expectations for every possible couple of indices.  In (7) we used again the definition of Z and in (8) we used the definition of the auto-correlation function of Z. In (9) we defined $\tau' = \tau + n - m $ to be the time difference between each couple of indices from $\bar{J}_t$ and $\bar{J}_{t+\tau}$. Note that $\tau' \in \tilde{J}_\tau$ where $\tilde{J}_\tau = \{ \tau - N + 1, \ldots, \tau + N - 1\}$. 

The calculation for the covariance $C_Y(\tau)$ follows the same steps as we did for $R_Y(\tau)$.
\end{proof}

\subsection{Proof of Lemma \ref{lemma:difftime}}
\label{app:proof_lemma:difftime}

\begin{proof}
Let $\bar{J}_t = \{\bar{j}_k\}_{k=1}^K \subset \{N, \ldots, 1\}$ and $\bar{J}_{t+\tau} = \{\bar{l}_k\}_{k=1}^K \subset \{N, \ldots, 1\}$
be subsets of K indices each. Let $J_t = \{j_k\}_{k=1}^K \subset \{t-N+1, \ldots, t\}$ and $J_{t+\tau} = \{l_k\}_{k=1}^K \subset \{t+\tau-N+1, \ldots, t + \tau\}$ be subsets of K time-steps each. The relations between these two subsets are: $\bar{j}_k = n \rightarrow j_k = t-n+1$ and $\bar{l}_k = m \rightarrow l_k = t+\tau-m+1$. 

We saw in Section \ref{app:prove_theorem:auto_corr_covariance} that we can move from these two subsets into the set of all possible differences $\tilde{J}_\tau$. Recall that we defined $\tau' = \tau + n - m $ to be the time difference between each couple of indices from $\bar{J}_t$ and $\bar{J}_{t+\tau}$. Note that $\tau' \in \tilde{J}_\tau$ where $\tilde{J}_\tau = \{ \tau - N + 1, \ldots, \tau + N - 1\}$. 

We now define some sets and multisets that will help us in the calculation of $P(\tau')$. 

\begin{definition}
Let $\mathbb{C}_{N,K}$ be the set of all possible permutations of choosing K samples out of N samples without replacement.
\end{definition}
 Observe that:
\[|\mathbb{C}_{N,K}| = \binom{N}{K} \]

\begin{definition}
Let $\mathbb{L}_{N,K}$ be the set of all possible tuples of two batches of K samples chosen from a set of N samples.
\end{definition}
 Observe that:
\[ |\mathbb{L}_{N,K}|  = \left(\binom{N}{K} \right)^2 \]

\begin{definition}
Let $\mathbb{M}_{N,K}$ be the multiset of all possible two-sample tuple generated for all possible couple of batches in the set $\mathbb{L}_{N,K}$.
Let $\bar{\mathbb{M}}_{N,K}$ be the unique set of  $\mathbb{M}_{N,K}$:
\end{definition}
Observe that:
\[|\mathbb{M}_{N,K}| = K^2 \cdot \left(\binom{N}{K} \right)^2 \]
\[ |\bar{\mathbb{M}}_{N,K}| = N^2 \]

\begin{definition}
Let $M(n,m)$ be the number of times the unique tuple $(n,m)$ appears in the multiset $\mathbb{M}_{N,K}$.
\end{definition} 
Observe that:
\[M(n,m) = \left(\binom{N-1}{K-1} \right)^2 \text{ for all } n \in \{1, \ldots, N\},   m \in \{\tau, \ldots, N+\tau\} \]

\begin{definition}
Let $D_{N,K,\tau'}$ be the number of unique sample tuples which have the time difference such that: $n-m=\tau'-\tau$ 
\end{definition}
Observe that: 
\[D_{N,K,\tau'} = N-|\tau'-\tau|\]

Here we consider the "unordered sampling without replacement" (described in Section \ref{section:setup}) for sampling $\bar{J}_t$ and $\bar{J}_{t+\tau}$ and we would like to calculate the probability distribution for each time difference $\tau'$, that is $P(\tau')$. We have total of $\NBATCH^2 \cdot \left(\binom{N}{\NBATCH}\right)^2$ such differences since we have $\binom{N}{\NBATCH}$ possible permutations for each batch and in each permutation we have $\NBATCH$ time elements. We saw in the definitions above that from all $\NBATCH^2 \cdot \left(\binom{N}{K} \right)^2$ possible two-sample couples we have $N^2$ unique sample couples, each of which has $\left(\binom{N-1}{K-1} \right)^2$ repetitions. From these $N^2$ couples, we have only $N-|\tau'-\tau|$ unique sample tuples $(n,m)$ that holds $n-m=\tau'-\tau$. We define $d = \tau'-\tau$, therefore $-N + 1 \le d \le N - 1$. We now can calculate $P(\tau')$: 
\begin{equation*}
        \begin{split}
            P(\tau') & \stackrel{}{=} \frac{ \left(\binom{N-1}{\NBATCH-1} \right)^2 \cdot (N-|\tau'-\tau|)}{\NBATCH^2 \cdot \left(\binom{N}{\NBATCH}\right)^2} \\
            & \stackrel{(1)}{=} \frac{ \left(\binom{N-1}{\NBATCH-1} \right)^2 \cdot (N-|d|)}{\NBATCH^2 \cdot \left(\binom{N}{\NBATCH}\right)^2} \\
            & \stackrel{(2)}{=} \frac{ (N-1)! \cdot  (N-1)!  \cdot \NBATCH! \cdot \NBATCH! \cdot (N-\NBATCH)! \cdot (N-\NBATCH)! \cdot (N-|d|)}{\NBATCH^2 \cdot (\NBATCH -1)! \cdot (\NBATCH -1)!  \cdot (N-\NBATCH)! \cdot (N-\NBATCH)! \cdot N! \cdot N! } \\
            & \stackrel{(3)}{=} \frac{ N-|d|}{N^2} \\
        \end{split}
    \end{equation*}
where in (1) we substitute $\tau'-\tau=d$. In (2) we wrote explicitly the binomial terms. In (3) we canceled similar elements in the denominator and numerator. Notice that this probability formula is relevant only for $\tau - N + 1\le \tau' \le \tau+N - 1$ and other values of $\tau'$ can not be reached form combining these two batches. Therefore, $P(\tau')=0$ for $\tau - N + 1 > \tau'$ and $\tau' > \tau+N - 1$.

Interestingly, this proof shows how parameter $\NBATCH$ is canceled out, meaning this time difference distribution is independent on $\NBATCH$. In addition, we can observe that the resulting distribution can be considered as a convolution of two rectangles, which represents the time limits of each batch and the uniform sampling, and the resulting convolution, a triangle which represents  $P(\tau')$.

\end{proof}

\subsection{Proof of Corollary \ref{corr:autocorr_covariance_binom_sampling}}
\label{app:proof_corr:autocorr_covariance_binom_sampling}

\begin{proof}
Combining Theorem \ref{theorem:auto_corr_covariance} and Lemma \ref{lemma:difftime} we get:
\begin{equation*}
    \begin{split}
        R_Y(\tau) & = \mathbb{E}_{\tau'} \left[ R_Z(\tau')  \right]  = \sum_{\tau'} P(\tau')  R_Z(\tau') \stackrel{1}{=} \sum_{d=-N+1}^{N-1} \frac{N-|d|}{N^2}  R_Z(d+\tau) 
    \end{split}
\end{equation*}
where in (1) we used $P(\tau')$ from Lemma \ref{lemma:difftime} and changed the variables: $d=\tau' - \tau$ for $\tau-N + 1\le \tau' \le \tau+N - 1$. 
Similar development can be done to  $C_Y(\tau)$. 
\end{proof}
\section{Proof of Theorem \ref{theorem:critic_conv}: Average reward and critic convergence }
\label{appendix:critic_conv}

\begin{proof}
Recall that our TD-error update Algorithm \ref{alg:LACRB} is defined as $\delta(\OO_j) = r(S_j, A_j) - \eta + \phi(S'_j)^\top w -  \phi(S_j)^\top w$,
where $\OO_j = \{S_j, A_j, r(S_j, A_j), S'_j\}$. In the critic update in Algorithm \ref{alg:LACRB} we use an empirical mean of TD-errors of several sampled observations, denoted as $\{\OO_j\}_{j\in J}$. Then, the critic update is defined as
\[w' = w + \alpha^w \frac{1}{\NBATCH} \sum_{j \in J}\delta(\OO_j) \phi (S_j).\]
where $J$ is a random subset of $\NBATCH$ samples from RB with size $N$.  Using the definition of the sampled random $K$ indices $\bar{J}$, instead of $J$, we can write the update as:
\[w' = w + \alpha^w \frac{1}{\NBATCH} \sum_{n \in \bar{J}} \delta(\OO_{t - n +1}) \phi (S_{t - n + 1}).\]

In this proof we follow the proof of Lemma 5 in \citet{bhatnagar2009natural}. Observe that the average reward and critic updates from Algorithm \ref{alg:LACRB} can be written as 
\begin{align}
\label{eq:eta_iteration}
\eta_{t+1} &= \eta_t + \alpha_t^{\eta} \left( F_t^{\eta} + M_{t+1}^\eta \right)\\
\label{eq:v_iteration}
    w_{t+1} &= v_t + \alpha_t^{w} \left( F_t^{w} + M_{t+1}^w \right),
\end{align}
where
\begin{equation*}
\label{eq:critic_Martingale_F}
\begin{split}
    F_t^{\eta} & \triangleq \mathbb{E}_{\bar{J}_t \sim \mathbb{C}_{N, K}, \{\OO_{t-n+1}\}_{n \in \bar{J}_t} \sim\text{RB}_t}  \left[ \frac{1}{\NBATCH} \sum_{n \in \bar{J}_t} r(S_{t-n+1},A_{t-n+1}) - \eta \middle| \mathcal{F}_t \right] \\
    M_{t+1}^\eta & \triangleq \left(\frac{1}{\NBATCH} \sum_{n \in \bar{J}_t} r(S_{t-n+1},A_{t-n+1}) - \eta_t \right) - F_t^{\eta} \\
    F_t^{w} & \triangleq \mathbb{E}_{\bar{J}_t \sim \mathbb{C}_{N, K}, \{\OO_{t-n+1}\}_{n \in \bar{J}_t} \sim\text{RB}_t} \left[  \frac{1}{\NBATCH} \sum_{n \in \bar{J}_t} \delta(\OO_{t - n + 1}) \phi(S_{t-n+1}) \middle| \mathcal{F}_t \right] \\
    M_{t+1}^{w} & \triangleq \frac{1}{\NBATCH} \sum_{n \in \bar{J}_t} \delta(\OO_{t-n+1}) \phi(S_{t-n+1}) - F_t^{w}
\end{split}
\end{equation*}
and $\mathcal{F}_t$ is a $\sigma$-algebra defined as
$\mathcal{F}_t \triangleq \{\eta_\tau, w_\tau, M_{\tau}^\eta, M_{\tau}^w: \tau \le t\}$. 

We use Theorem 2.2 of \citet{borkar2000ode} to prove convergence of these iterates. Briefly, this theorem
states that given an iteration as in \eqref{eq:eta_iteration} and  \eqref{eq:v_iteration}, these iterations are bounded w.p.1 if

\begin{assumption}
\label{ass:borkar_ass_v}
\begin{enumerate}
\item $F_t^{\eta}$ and $F_t^{w}$  are Lipschitz, the functions $F_\infty(\eta) = \lim_{\sigma \rightarrow \infty} F^{\eta} (\sigma \eta)/\sigma$ and $F_\infty (w) = \lim_{\sigma \rightarrow \infty} F^{w} (\sigma w )/\sigma$ are Lipschitz, and $F_\infty(\eta)$ and $F_\infty(w)$ are asymptotically stable in the origin.\\
\item The sequences $M_{t+1}^{\eta}$ and $M_{t+1}^{w}$  are  martingale difference noises and for some $C_0^\eta$, $C_0^w$
\[\mathbb{E}\left[ (M_{t+1}^{\eta})^2 \middle| \mathcal{F}_t \right] \le C_0^\eta (1 + \| \eta_t \|^2)\]
\[\mathbb{E}\left[ (M_{t+1}^{w})^2 \middle| \mathcal{F}_t \right] \le C_0^w (1 + \| w_t \|^2).\]
\end{enumerate}
\end{assumption}
We begin with  the average reward update in \eqref{eq:eta_iteration}. The ODE describing its asymptotic behavior corresponds to 
\begin{equation}
\label{eq:eta_dot}
    \dot \eta = \mathbb{E}_{\bar{J} \sim \mathbb{C}_{N, K}, \{\OO_{t-n+1}\}_{n \in \bar{J}} \sim\text{RB}}  \left[ \frac{1}{\NBATCH} \sum_{n \in \bar{J}_t} r(S_{t-n+1},A_{t-n+1}) - \eta \right] \triangleq F^\eta.
\end{equation}

$F^\eta$ is Lipschitz continuous in $\eta$. The function $F_\infty (\eta)$ exists and satisfies $F_\infty (\eta) = -\eta$. The origin is an asymptotically stable equilibrium for the ODE $\dot \eta = F_\infty (\eta)$ and the related Lyapunov function is given by $\eta^2/2$.

For the critic update, consider the ODE 
\[\dot{w} = \mathbb{E}_{\bar{J} \sim \mathbb{C}_{N, K}, \{\OO_{t-n+1}\}_{n \in \bar{J}} \sim\text{RB}} \left[  \frac{1}{\NBATCH} \sum_{n \in \bar{J}} \delta(\OO_{t - n + 1}) \phi(S_{t-n+1})  \right] \triangleq F^w\]
In Lemma \ref{lemma:expected_delta} we show that this ODE can be written as
\begin{equation}
\label{eq:ode_v2}
 \dot{w}
    = \Phi^\top C_\theta \Phi w +  \Phi^\top b_\theta,
\end{equation}
where $C_\theta$ and $b_\theta$ are defined in \eqref{eq:def_A_b_of_TD_infty}. $F^w$ is Lipschitz continuous in $w$ and $F_\infty (w)$ exists and satisfies $F_\infty (w) = \Phi^\top C_\theta \Phi w$. 
Consider the system
\begin{equation}
\label{eq:ode_v_infty2}
    \dot{w} =F_\infty (w)
\end{equation}
In assumption \ref{ass:independent_features} we assume that $\Phi w \neq e$ for every $w \in \mathbb{R}^d$. Therefore, the only asymptotically stable equilibrium for \eqref{eq:ode_v_infty2} is the origin (see the explanation in the proof of Lemma 5 in \citet{bhatnagar2009natural}). Therefore, for all $t \ge 0$
\[\mathbb{E}\left[ (M_{t+1}^{\eta})^2 \middle| \mathcal{F}_t \right] \le C_0^\eta (1 + \| \eta_t \|^2 + \| w_t \|^2)\]
\[\mathbb{E}\left[ (M_{t+1}^{w})^2 \middle| \mathcal{F}_t \right] \le C_0^w (1 + \| \eta_t \|^2 + \| w_t \|^2)\]
for some $C_0^\eta, C_0^w < \infty$. $M_{t}^{\eta}$ can be directly seen to be uniformly bounded almost surely. Thus, Assumptions (A1) and (A2) of \citet{borkar2000ode} are satisfied for the average reward, TD-error, and critic updates. From Theorem 2.1 of \citet{borkar2000ode},  the average reward, TD-error, and critic iterates are uniformly bounded with probability one. Note that when $t \rightarrow \infty$, \eqref{eq:eta_dot} has $\eta_\theta$ defined as in \eqref{eq:actor_loss} as its unique globally asymptotically stable equilibrium with $V_2(\eta) = (\eta - \eta_\theta)^2 $ serving as the associated Lyapunov function.

Next, suppose that $w = w^\pi$ is a solution to the system $\Phi^\top C_\theta  \Phi w  = 0.$ Under Assumption \ref{ass:independent_features}, using the same arguments as in the proof of Lemma 5 in \citet{bhatnagar2009natural}, $w^\pi$ is the unique globally asymptotically stable equilibrium of the ODE \eqref{eq:ode_v2}. Assumption \ref{ass:borkar_ass_v} is now verified and under Assumption \ref{ass:step_size}, the claim follows from Theorem 2.2, pp. 450 of \cite{borkar2000ode}.
\end{proof}
\subsection{Proof of Lemma \ref{lemma:expected_delta}}
\label{app:proof_lemma_expected_delta}
\begin{proof}
We compute the expectation of the critic update with linear function approximation according to Algorithm \ref{alg:LACRB}. In this proof, we focus on the "Unordered sampling without replacement" strategy for sampling batch of $K$ transitions from the replay buffer (see Section \ref{sec:smampling_method} for this strategy probability distribution). Recall that $n$ is a position in the RB and it corresponds to transition $\OO_{t-n+1}=(S_{t-n+1}, A_{t-n+1}, S'_{t-n+1})$. We will use the notation of $\bar{J} \subset \{1, \ldots, n, \ldots, N\}$ to refer the $K$ indices sampled batches. In addition we will use the following observations:
\begin{align}
    \label{eq:n_prob_in_J} P(n|\bar{J}, n \in \bar{J} ) & = \frac{1}{K}, \quad  P(n|\bar{J},n \notin \bar{J}) = 0 \\
     \nonumber P(n \in \bar{J}) & = \frac{K}{N}, \quad P(n \notin \bar{J}) = 1 - \frac{K}{N} \\
     \label{eq:n_given_J} P(n|\bar{J}) & = P(n \in \bar{J}) \cdot P(n|\bar{J}, n \in \bar{J}) + P(n \notin \bar{J}) \cdot P(n|\bar{J},n \notin \bar{J}) = \frac{K}{N} \frac{1}{K} + 0 = \frac{1}{N} \\
     \label{eq:bar_J_prob} P(\bar{J}) & = \frac{1}{\binom{N}{K}} \\
     \label{eq:C_n_K_size}| \mathbb{C}_{N,K} | & = \binom{N}{K}
\end{align}
Now we can compute the desired expectation:
\begin{equation}
\label{eq:grand_expectation}
\begin{split}
    & \mathbb{E}_{\bar{J}_t \sim \mathbb{C}_{N, K}, \{\OO_{t-n+1}\}_{n \in \bar{J}_t} \sim\text{RB}_t} \left[  \frac{1}{\NBATCH} \sum_{n \in \bar{J}_t} \delta(\OO_{t - n + 1}) \phi(S_{t-n+1}) \right] \\
    & \stackrel{}{=}
    \mathbb{E}_{\bar{J}_t \sim \mathbb{C}_{N, K}} 
     \left[ \mathbb{E}_{\{\OO_{t-n+1}\}_{n \in \bar{J}_t} \sim \text{RB}_t}  \left[ \frac{1}{\NBATCH} \sum_{n \in \bar{J}_t} \delta(\OO_{t - n + 1}) \phi(S_{t-n+1})  \middle| \bar{J}_t\right] \right] \\
     & \stackrel{\eqref{eq:n_prob_in_J}}{=}
    \mathbb{E}_{\bar{J}_t \sim \mathbb{C}_{N, K}} 
     \left[ \mathbb{E}_{\{\OO_{t-n+1}\}_{n \in \bar{J}_t} \sim \text{RB}_t}  \left[ \mathbb{E}_{n \sim \bar{J}_t} \left[ \delta(\OO_{t - n + 1}) \phi(S_{t-n+1}) \right] \middle| \bar{J}_t\right] \right] \\
     & \stackrel{1}{=}
    \mathbb{E}_{\bar{J}_t \sim \mathbb{C}_{N, K}} 
     \left[  \mathbb{E}_{n \sim \bar{J}_t} \left[ \mathbb{E}_{\OO_{t-n+1}} \left[  \delta(\OO_{t - n + 1}) \phi(S_{t-n+1}) \right] \right] \middle| \bar{J}_t \right] \\
    & \stackrel{2}{=}  \sum_{\bar{J}_t \in \mathbb{C}_{N, K}} P(\bar{J}_t)  \sum_{n=1}^N P(n|\bar{J}_t) \mathbb{E}_{\OO_{t-n+1}}     
    \left[\delta(\OO_{t-n+1}) \phi(S_{t-n+1}) \right]\\
    & \stackrel{\eqref{eq:n_given_J}, \eqref{eq:bar_J_prob}}{=}  \sum_{\bar{J}_t\in \mathbb{C}_{N, K}} \frac{1}{\binom{N}{K}} \sum_{n=1}^N \frac{1}{N} \mathbb{E}_{\OO_{t-n+1}}   
    \left[\delta(\OO_{t-n+1}) \phi(S_{t-n+1}) \right]\\
    & \stackrel{\eqref{eq:C_n_K_size}}{=} \frac{1}{N} \sum_{n=1}^N  \mathbb{E}_{\OO_{t-n+1}}     
    \left[\delta(\OO_{t-n+1}) \phi(S_{t-n+1}) \right]\\
    & \stackrel{3}{=} \frac{1}{N} \sum_{n=1}^N  \mathbb{E}_{S_{t-n+1}, A_{t-n+1}, S'_{t-n+1}}     
    \left[\left( r(S_{t-n+1}, A_{t-n+1}) - \eta + \phi(S'_{t-n+1})^\top w -  \phi(S_{t-n+1})^\top w \right) \phi(S_{t-n+1})  \right]
\end{split}
\end{equation}

where  in (1) we are left with the marginal expectations for each observation, in (2) we wrote expectations explicitly and  in (3) we used the definition of the TD-error in \eqref{eq:TD_error}. 

Next, for time $t - n + 1$ where $1 \le n \le N$, we define the induced MC with a corresponding policy parameter $\theta_{t - n +1}$. For this parameter, we denote the corresponding state distribution vector $\rho_{t - n + 1}$ and a transition matrix $P_{t - n + 1}$ (both induced by the policy $\pi_{\theta_{t - n + 1}}$. In addition, we define the following diagonal matrix $D_{t - n + 1}\triangleq \textrm{diag}(\rho_{t - n +1})$. Similarly to  \cite{bertsekas1996neuro} Lemma 6.5, pp.298, we can substitute the inner expectation 
\begin{equation}
\label{eq:inner}
\begin{split}
   & \ \mathbb{E}_{S_{t-n+1}, A_{t-n+1}, S'_{t-n+1}}     
    \left[\left( r(S_{t-n+1}, A_{t-n+1}) - \eta + \phi(S'_{t-n+1})^\top w -  \phi(S_{t-n+1})^\top w \right) \phi(S_{t-n+1})  \right] \\
    & =  \Phi^\top D_{t - n +1} \left(P_{t - n + 1} - I \right) \Phi w + \Phi^\top D_{t - n + 1} (r_{t - n +1}- \eta_{\theta}  e), 
\end{split}
\end{equation}
where $I$ is the $|\mathcal{S}| \times |\mathcal{S}|$ identity matrix, $e$ in $|\mathcal{S}| \times 1$ vector of ones and $r_{t - n + 1}$ is a $|\mathcal{S}| \times 1$ vector defined as $r_{t - n + 1}(s) = \sum_{a} \pi_{\theta_{t - n + 1}} (A|S) r(S,A)$. Combining equations \eqref{eq:def_A_b_of_TD},  \eqref{eq:grand_expectation} and \eqref{eq:inner} yields
\begin{equation}
\begin{split}
    \label{eq:k_envs_td_way1}
    \frac{1}{N} \sum_{n=1}^N 
    \left(
    \Phi^\top D_{t - n + 1} \left(P_{t - n +1} - I\right) \Phi w + \Phi^\top D_{t - n + 1} (r_{t - n +1}-\eta_{\theta} e) \right)= \Phi^\top C_t \Phi w +  \Phi^\top b_t,
\end{split}
\end{equation}
In the limit, $t \rightarrow \infty$ and $\rho_{t - n +1} \rightarrow \mu_{\theta}$ for all index $n$. Using $C_\theta$ and $b_\theta$ defined in \eqref{eq:def_A_b_of_TD_infty}, \eqref{eq:grand_expectation} can be expressed as 
\begin{equation}
\begin{split}
    \mathbb{E}_{\bar{J}_t \sim \mathbb{C}_{N, K}, \{\OO_{t-n+1}\}_{n \in \bar{J}_t} \sim\text{RB}_t} \left[  \frac{1}{\NBATCH} \sum_{n \in \bar{J}_t} \delta(\OO_{t - n + 1}) \phi(S_{t-n+1}) \right] 
    & =
    \Phi^\top C_\theta \Phi w +  \Phi^\top b_\theta.
\end{split}
\end{equation}

\end{proof}
\newpage
\section{Proof of Theorem \ref{theorem:actor_conv}: Actor convergence}
\label{appendix:actor_conv}

\begin{proof}
Recall that our TD-error update in Algorithm \ref{alg:LACRB} is defined as $\delta(\OO_j) = r(S_j, A_j) - \eta + \phi(S'_j)^\top w -  \phi(S_j)^\top w$,
where $\OO_j = \{S_j, A_j, r(S_j, A_j), S'_j\}$. In the actor update in Algorithm \ref{alg:LACRB} we use an empirical mean of TD-errors of several sampled observations, denoted as $\{\OO_j\}_{j\in J}$. Then, the actor update is defined as
\[\theta' = \Gamma \left(\theta - \alpha^\theta \frac{1}{\NBATCH} \sum_{j \in J}\delta(\OO_j)  \nabla \log \pi_\theta (A_j |S_j) \right).\]
where $J$ is a random subset of $\NBATCH$ samples from RB with size $N$. Using the definition of the sampled random $K$ indices $\bar{J}$, instead of $J$, we can write the update as:
\[\theta' = \Gamma \left(\theta - \alpha^\theta \frac{1}{\NBATCH} \sum_{n \in \bar{J}}\delta(\OO_{t-n+1})  \nabla \log \pi_\theta (A_{t-n+1} |S_{t-n+1}) \right).\]

In this proof we follow the proof of Theorem 2 in \citet{bhatnagar2009natural}. Let $\OO = \{S, A, S'\}$ and let $\delta^\pi(\OO) = r(S,A) - \eta + \phi(S')^\top w^\pi - \phi(S)^\top w^\pi$, where $w^\pi$ is the convergent parameter of the critic recursion with probability one (see its definition in the proof for Theorem \ref{theorem:critic_conv}). Observe that the actor parameter update from Algorithm  \ref{alg:LACRB} can be written as 
\begin{align*}
    \theta_{t+1}  &= \Gamma \Big( \theta_t - \alpha^{\theta}_t \big( \delta(\OO) \nabla_\theta \log \pi_{\theta} (A|S)  + F_t^{\theta} -  F_t^{\theta} + N_t^{\theta_t} - N_t^{\theta_t} \big) \Big) \\
    & = \Gamma \Big( \theta_t -  \alpha^{\theta}_t \big( M_{t+1}^{\theta} + ( F_{t}^{\theta} - N_{t}^{\theta_t} ) +  N_{t}^{\theta_t} \big) \Big)
\end{align*}
where
\begin{equation*}
\label{eq:actor_Martingale_F}
\begin{split}
    F_t^{\theta} & \triangleq \mathbb{E}_{\bar{J}_t \sim \mathbb{C}_{N, K}, \{\OO_{t-n+1}\}_{n \in \bar{J}_t} \sim\text{RB}_t} \left[  \frac{1}{\NBATCH} \sum_{n \in \bar{J}_t} \delta(\OO_{t - n + 1}) \nabla_\theta \log \pi_{\theta} (A_{t - n + 1}|S_{t - n + 1}) \middle| \mathcal{F}_t \right] \\
    M_{t+1}^{\theta} & \triangleq \frac{1}{\NBATCH} \sum_{n \in \bar{J}_t} \delta(\OO_{t - n + 1}) \nabla_\theta \log \pi_{\theta} (A_{t - n + 1}|S_{t - n + 1})  - F_t^{\theta}\\
    N_t^{\theta} & \triangleq \mathbb{E}_{\bar{J}_t \sim \mathbb{C}_{N, K}, \{\OO_{t-n+1}\}_{n \in \bar{J}_t} \sim\text{RB}_t} \left[  \frac{1}{\NBATCH} \sum_{n \in \bar{J}_t} \delta^{\pi_\theta}(\OO_{t - n + 1}) \nabla_\theta \log \pi_{\theta} (A_{t - n + 1}|S_{t - n + 1}) \middle| \mathcal{F}_t \right]
\end{split}
\end{equation*}

and $\mathcal{F}_t$ is a $\sigma$-algebra defined as
$\mathcal{F}_t \triangleq \{\eta_\tau, w_\tau, \theta_\tau, M_{\tau}^\eta, M_{\tau}^w, M_{\tau}^\theta: \tau \le t\}$. 

Since the critic converges along the faster timescale, from Theorem \ref{theorem:critic_conv} it follows that $F_{t}^{\theta} - N_{t}^{\theta_t} = o(1)$. Now, let
\[M_2(t) = \sum_{r=0}^{t-1} \alpha^{\theta}_r M_{r+1}^\theta, t \ge 1.\]
The quantities $\delta (\OO)$ can be seen to be uniformly bounded since from the proof in Theorem \ref{theorem:critic_conv}, $\{ \eta_{t} \}$ and $\{w_t \}$ are bounded sequences. Therefore, using Assumption \ref{ass:step_size}, $\{ M_2(t)\}$ is a convergent martingale sequence \cite{bhatnagar2004simultaneous}. 

Consider the actor update along the slower timescale corresponding to $\alpha^\theta_t$  in Algorithm \ref{alg:LACRB}. Let $w(\cdot)$ be a vector field on a set $\Theta$. Define another vector field: $\hat \Gamma \big(w(y) \big) = \lim_{0 < \eta \rightarrow 0} \Big( \frac{\Gamma\big(y + \eta w(y) \big) - y}{\eta}\Big)$. In case this limit is not unique, we let $\hat \Gamma \big(w(y) \big)$ be the set of all possible limit points (see pp. 191 of \cite{kushner2012stochastic}). Consider now the ODE
\begin{align}
\label{eq:ode_theta_bias}
    \dot \theta & = \hat \Gamma \left( - \mathbb{E}_{\bar{J}_t \sim \mathbb{C}_{N, K}, \{\OO_{t-n+1}\}_{n \in \bar{J}_t} \sim\text{RB}_t} \left[  \frac{1}{\NBATCH} \sum_{n \in \bar{J}_t} \delta^{\pi_\theta}(\OO_{t - n + 1}) \nabla_\theta \log \pi_{\theta} (A_{t - n + 1}|S_{t - n + 1}) \right] \right) 
\end{align}
Substituting the result from Lemma \ref{lemma:expected_delta_grad_log}, the above ODE is analogous to 
\begin{equation}
\label{eq:ode_theta_bias2}
    \dot \theta = \hat \Gamma (- \nabla_\theta \eta_\theta + \xi^{\pi_\theta}) = \hat \Gamma \big( - N_t^\theta \big)
\end{equation}

where $\xi^{\pi_\theta} = \sum_{S} \mu_{\theta}  (S) \Big(  \phi(S)^\top \nabla_\theta w^{\pi_\theta} -  \nabla_\theta \bar{V}^{\pi_{\theta}}(S) \Big) $. Consider also an associated ODE:
\begin{equation}
\label{eq:ode_theta}
    \dot \theta  = \hat \Gamma \big(- \nabla_\theta \eta_\theta \big) 
\end{equation}

We now show that $h_1(\theta_t) \triangleq - N_t^{\theta_t}$ is Lipschitz continuous. Here $w^{\pi_{\theta_t}}$ corresponds to the weight vector to which the critic update converges along the faster timescale when the corresponding policy is $\pi_{\theta_t}$ (see Theorem \ref{theorem:critic_conv}). Note that $\mu_{\theta}(S), S \in \mathcal{S}$ is continuously differentiable in $\theta$ and have bounded derivatives. Also, $\bar{\eta}_{\theta_t}$ is continuously differentiable as well and has bounded derivative as can also be seen from \eqref{eq:actor_loss}. Further, $w^{\pi_{\theta_t}}$ can be seen to be continuously differentiable with bounded derivatives. Finally, $\nabla^2 \pi_{\theta_t}(A|S)$ exists and is bounded. Thus $ h_1(\theta_t)$ is a Lipschitz continuous function and the ODE (\ref{eq:ode_theta_bias}) is well posed.

Let $\mathcal{Z}$ denote the set of asymptotically stable equilibria of (\ref{eq:ode_theta}) i.e., the local minima of $\eta_\theta$, and let $\mathcal{Z}^\epsilon$ be the $\epsilon$-neighborhood of $\mathcal{Z}$. To complete the proof, we are left to show that as $\sup_\theta \| \xi^{\pi_\theta}\| \rightarrow 0 $ (viz. $\delta \rightarrow 0 $), the trajectories of (\ref{eq:ode_theta_bias2}) converge to those of (\ref{eq:ode_theta}) uniformly on compacts for the same initial condition in both. This claim follows the same arguments as in the proof of Theorem 2 in \citet{bhatnagar2009natural}.
\end{proof}

\subsection{Proof of Lemma \ref{lemma:expected_delta_grad_log}}
\label{app:proof of actor lemma}

\begin{proof}
We compute the required expectation with linear function approximation according to Algorithm \ref{alg:LACRB}. Following the same steps when proving the expectation for the critic in Section \ref{app:proof_lemma_expected_delta}, we have:
{\small
\begin{equation*}
\begin{split}
    & \mathbb{E}_{\bar{J}_t \sim \mathbb{C}_{N, K}, \{\OO_{t-n+1}\}_{n \in \bar{J}_t} \sim\text{RB}_t} \left[  \frac{1}{\NBATCH} \sum_{n \in \bar{J}_t} \delta^{\pi_\theta}(\OO_{t - n + 1}) \nabla_\theta \log \pi_{\theta} (A_{t - n + 1}|S_{t - n + 1}) \right] \\
    & \stackrel{}{=} \frac{1}{N} \sum_{n=1}^N  \mathbb{E}_{S_{t-n+1}, A_{t-n+1}, S'_{t-n+1}}     
    \left[\left( r(S_{t-n+1}, A_{t-n+1}) - \eta + \phi(S'_{t-n+1})^\top w -  \phi(S_{t-n+1})^\top w \right) \nabla_\theta \log \pi_{\theta} (A_{t - n + 1}|S_{t - n + 1})  \right]
\end{split}
\end{equation*}
}
Recall the definition of the  state distribution vector $\rho_{t - n +1}$ in  Section \ref{subsec:Algorithm_1_asymptotic_convergence}.  In the limit, $t \rightarrow \infty$ and $\rho_{t-n +1} \rightarrow \mu_{\theta}$ for all index $n$, then:
\begin{equation*}
\begin{split}
    & \mathbb{E}_{\bar{J}_t \sim \mathbb{C}_{N, K}, \{\OO_{t-n+1}\}_{n \in \bar{J}_t} \sim\text{RB}_t} \left[  \frac{1}{\NBATCH} \sum_{n \in \bar{J}_t} \delta^{\pi_\theta}(\OO_{t - n + 1}) \nabla_\theta \log \pi_{\theta} (A_{t - n + 1}|S_{t - n + 1}) \right] \\
    & = \sum_{S \in \mathcal{S}} \mu_{\theta} (S) \sum_{A \in \mathcal{A}} \pi_\theta (A|S)  \left(  r(S,A) - \eta_\theta + \sum_{S' \in \mathcal{S}} P(S'|S, A) \phi(S')^\top w^{\pi_\theta} - \phi(S)^\top w^{\pi_\theta} \right) \nabla_\theta \log \pi_{\theta} (A|S)  \\
\end{split}
\end{equation*}

We define now the following term:
\begin{equation}
\label{eq:expected_v_pi}
    \begin{split}
        \bar{V}^{\pi_{\theta}}(S) &= \sum_{A \in \mathcal{A}}  \pi_{\theta} (A|S) \bar{Q}^{\pi_{\theta}} (S, A) 
        =\sum_{A \in \mathcal{A}} \pi_{\theta} (A|S) \left( r(S,A) - \eta_{\theta} + \sum_{S' \in \mathcal{S}} P (S'|S, A) \phi (S')^\top w^{\pi_\theta} \right),
    \end{split}
\end{equation}
where $ \bar{V}^{\pi_{\theta}}(S)$ and $ \bar{Q}^{\pi_{\theta}}(S, A)$ correspond to policy $\pi_{\theta}$. Note that here, the convergent critic parameter $w^{\pi_\theta}$ is used. Let's look at the gradient of \eqref{eq:expected_v_pi}:

\begin{equation*}
    \begin{split}
        \nabla_\theta \bar{V}^{\pi_{\theta}}(S) & = \nabla_\theta \left(\sum_{A \in \mathcal{A}}  \pi_{\theta} (A|S) \bar{Q}^{\pi_{\theta}} (S, A) \right) \\
        & = \sum_{A \in \mathcal{A}} \nabla_\theta \pi_{\theta} (A|S) \left( r(S,A) - \eta_{\theta} + \sum_{S' \in \mathcal{S}} P (S'|S, A) \phi (S')^\top w^{\pi_\theta} \right) \\
        & + \sum_{A \in \mathcal{A}} \pi_{\theta} (A|S) \left( - \nabla_\theta \eta_{\theta} + \sum_{S' \in \mathcal{S}} P (S'|S, A) \phi (S')^\top \nabla_\theta w^{\pi_\theta} \right) \\
        & = \sum_{A \in \mathcal{A}} \nabla_\theta \pi_{\theta} (A|S) \left( r(S,A) - \eta_{\theta} + \sum_{S' \in \mathcal{S}} P (S'|S, A) \phi (S')^\top w^{\pi_\theta} \right) \\
        & - \nabla_\theta \eta_{\theta}  + \sum_{A \in \mathcal{A}} \pi_{\theta} (A|S) \sum_{S' \in \mathcal{S}} P(S'|S, A) \phi (S')^\top \nabla_\theta w^{\pi_\theta} 
    \end{split}
\end{equation*}

Summing both sides over the stationary distribution $\mu_{\theta}$
\begin{equation*}
    \begin{split}
         \sum_{S} \mu_{\theta} (S) \nabla_\theta \bar{V}^{\pi_{\theta}}(S) 
        & = \sum_{S} \mu_{\theta} (S) 
        \sum_{A \in \mathcal{A}} \nabla_\theta \pi_{\theta} (A|S) \left( r(S, A) - \eta_{\theta} + \sum_{S' \in \mathcal{S}} P (S'|S, A) \phi (S')^\top w^{\pi_\theta} \right) \\
        & +  \sum_{S} \mu_{\theta} (S) \left( - \nabla_\theta \eta_{\theta}  + \sum_{A \in \mathcal{A}} \pi_{\theta} (A|S)  \sum_{S' \in \mathcal{S}} P (S'|S, A) \phi (S')^\top \nabla_\theta w^{\pi_\theta} \right) \\
        & = \mathbb{E}_{\bar{J}_t \sim \mathbb{C}_{N, K}, \{\OO_{t-n+1}\}_{n \in \bar{J}_t} \sim\text{RB}_t} \left[  \frac{1}{\NBATCH} \sum_{n \in \bar{J}_t} \delta^{\pi_\theta}(\OO_{t - n + 1}) \nabla_\theta \log \pi_{\theta} (A_{t - n + 1}|S_{t - n + 1}) \right] \\
        & - \nabla_\theta \eta_\theta  + \sum_{S} \mu_{\theta} (S) \sum_{A \in \mathcal{A}} \pi_{\theta} (A|S) \sum_{S' \in \mathcal{S}} P (S'|S, A) \phi (S')^\top \nabla_\theta w^{\pi_\theta} 
    \end{split}
\end{equation*}
Then:
\begin{equation*}
    \begin{split}
        \nabla_\theta \eta_\theta &= \mathbb{E}_{\bar{J}_t \sim \mathbb{C}_{N, K}, \{\OO_{t-n+1}\}_{n \in \bar{J}_t} \sim\text{RB}_t} \left[  \frac{1}{\NBATCH} \sum_{n \in \bar{J}_t} \delta^{\pi_\theta}(\OO_{t - n + 1}) \nabla_\theta \log \pi_{\theta} (A_{t - n + 1}|S_{t - n + 1}) \right] \\
        & + \sum_{S} \mu_{\theta} (S) \left( \sum_{A \in \mathcal{A}} \pi_{\theta} (A|S)  \sum_{S' \in \mathcal{S}} P (S'|S, A) \phi (S')^\top \nabla_\theta v^{\pi_\theta}  - \nabla_\theta \bar{V}^{\pi_{\theta}}(S)\right).
    \end{split}
\end{equation*}
Since $\mu_{\theta}$ is a stationary distribution, 
\begin{equation*}
    \begin{split}
        \sum_{S}  \mu_{\theta} (S) \sum_{A \in \mathcal{A}} \pi_{\theta} (A|S)  \sum_{S' \in \mathcal{S}} P (S'|S, A) \phi (S')^\top \nabla_\theta w^{\pi_\theta}  
        & = \sum_{S}  \mu_{\theta} (S)  \sum_{S' \in \mathcal{S}} P_{\theta} (S'|S) \phi (S')^\top \nabla_\theta w^{\pi_\theta} \\
        & = \sum_{S'} \sum_{S}  \mu_{\theta} (S)   P_{\theta} (S'|S) \phi (S')^\top \nabla_\theta w^{\pi_\theta} \\
        & = \sum_{S'} \mu_{\theta} (S') \phi (S')^\top \nabla_\theta w^{\pi_\theta},
    \end{split}
\end{equation*}
Then,
\begin{equation*}
    \begin{split}
        \nabla_\theta \eta_\theta & =
        \mathbb{E}_{\bar{J}_t \sim \mathbb{C}_{N, K}, \{\OO_{t-n+1}\}_{n \in \bar{J}_t} \sim\text{RB}_t} \left[  \frac{1}{\NBATCH} \sum_{n \in \bar{J}_t} \delta^{\pi_\theta}(\OO_{t - n + 1}) \nabla_\theta \log \pi_{\theta} (A_{t - n + 1}|S_{t - n + 1}) \right] \\
        & + \sum_{S} \mu_{\theta}  (S) \Big(  \phi(S)^\top \nabla_\theta w^{\pi_\theta} -  \nabla_\theta \bar{V}^{\pi_{\theta}}(S) \Big)
    \end{split}
\end{equation*}
The result follows immediately.
\end{proof}

\end{document}